\pdfoutput=1  %
\PassOptionsToPackage{table}{xcolor}  %
\documentclass[sigconf, 9pt, nonacm]{acmart}

\usepackage{siunitx}
\usepackage{subfig}
\usepackage{array}
\usepackage{tabularx}
\usepackage{booktabs}
\usepackage{multirow}
\usepackage{makecell}
\usepackage{threeparttable}
\usepackage{bm}
\usepackage{tikz}
\usetikzlibrary{positioning,arrows.meta,fit,backgrounds,calc}
\usepackage{flafter}                %
\usepackage[capitalise]{cleveref}   %

\setcopyright{none}
\renewcommand\footnotetextcopyrightpermission[1]{}

\newcommand{\vect}[1]{\mathbf{#1}}
\newcommand{\mat}[1]{\mathbf{#1}}
\def\transp{\mathsf{T}}            %
\newcommand{\inv}{^{-1}}
\DeclareMathOperator*{\med}{med}

\newcommand{\egovel}{\vect{v}^R}
\newcommand{\dopp}{d}
\newcommand{\uvec}[1]{\hat{\vect{#1}}}
\def\state{\vect{x}}               %
\newcommand{\errorstate}{\delta\vect{x}}
\newcommand{\noise}{\vect{n}}
\newcommand{\cov}{\mat{P}}
\newcommand{\meas}{\vect{z}}
\newcommand{\Hmat}{\mat{H}}

\newcommand{\Qmat}{\mat{Q}}

\newcommand{\Fmat}{\mat{F}}
\newcommand{\Gmat}{\mat{G}}
\newcommand{\Identity}{\mat{I}}
\newcommand{\ptcloudonboard}{\texttt{ptcloud\_\allowbreak onboard}}
\newcommand{\ptclouddspfft}{\texttt{ptcloud\_\allowbreak dsp\_\allowbreak fft}}
\newcommand{\ptclouddspcapon}{\texttt{ptcloud\_\allowbreak dsp\_\allowbreak capon}}
\newcommand{\alignoriginflag}{\texttt{--align\_\allowbreak origin}}

\definecolor{TolBrightBlue}{HTML}{4477AA}
\definecolor{TolBrightCyan}{HTML}{66CCEE}
\definecolor{TolBrightGreen}{HTML}{228833}
\definecolor{TolBrightYellow}{HTML}{CCBB44}
\definecolor{TolBrightRed}{HTML}{EE6677}
\definecolor{TolBrightPurple}{HTML}{AA3377}
\definecolor{TolBrightGrey}{HTML}{BBBBBB}

\definecolor{TolLightBlue}{HTML}{77AADD}
\definecolor{TolLightCyan}{HTML}{99DDFF}
\definecolor{TolMint}{HTML}{44BB99}
\definecolor{TolPear}{HTML}{BBCC33}
\definecolor{TolOliveLight}{HTML}{AAAA00}
\definecolor{TolLightYellow}{HTML}{EEDD88}
\definecolor{OIGreen}{HTML}{009E73}   %

\definecolor{Grey}{gray}{0.92}

\def\BibTeX{{\rm B\kern-.05em{\sc i\kern-.025em b}\kern-.08em
    T\kern-.1667em\lower.2ex\hbox{E}\kern-.125emX}}

\begin{document}

\title{Dense Soft Weighting for Radar Ego-Velocity Estimation}

\author{Atar Babgei, Chenyu Zhao, Michael Breza, Julie A. McCann}
\affiliation{%
  \institution{Imperial College London}
  \country{}
}
\email{{a.babgei22, chenyu.zhao25, michael.breza04, j.mccann}@imperial.ac.uk}

\renewcommand{\shortauthors}{Babgei et al.}

\begin{abstract}
Sensing ego-velocity estimation is fundamental to state estimation in
visually degraded environments, where camera- and LiDAR-based pipelines can
become unreliable. Millimetre-wave radar is well suited to these conditions
because it provides direct Doppler velocity sensing and remains robust to poor
illumination, textureless scenes, and airborne particulates. However,
conventional radar ego-velocity pipelines typically apply constant false alarm
rate (CFAR) thresholding to convert dense radar spectra into sparse point
clouds, prematurely discarding sub-threshold returns that may still retain
useful Doppler motion cues. We present \emph{Dense Soft Weighting}, an analytic radar front-end that maps every range-Doppler cell to a continuous confidence metric rather than
enforcing a binary detection threshold. Ego-velocity is then estimated using a
deterministic robust weighted least-squares formulation, while the same
weighted measurements provide a closed-form, measurement-derived velocity
covariance for integration with a shared inertial back-end. The method requires
no platform-specific training data or learning-based uncertainty model,
supporting transfer across single-chip radar configurations. Across two public
datasets and one self-collected dataset, Dense Soft Weighting reduces mean
absolute pose error by 31--45\% relative to the strongest CFAR point-cloud
baseline under an identical inertial back-end, while running in real time on
embedded hardware.
\end{abstract}


\maketitle

\begin{figure}[t]
    \centering
    \includegraphics[width=\columnwidth]{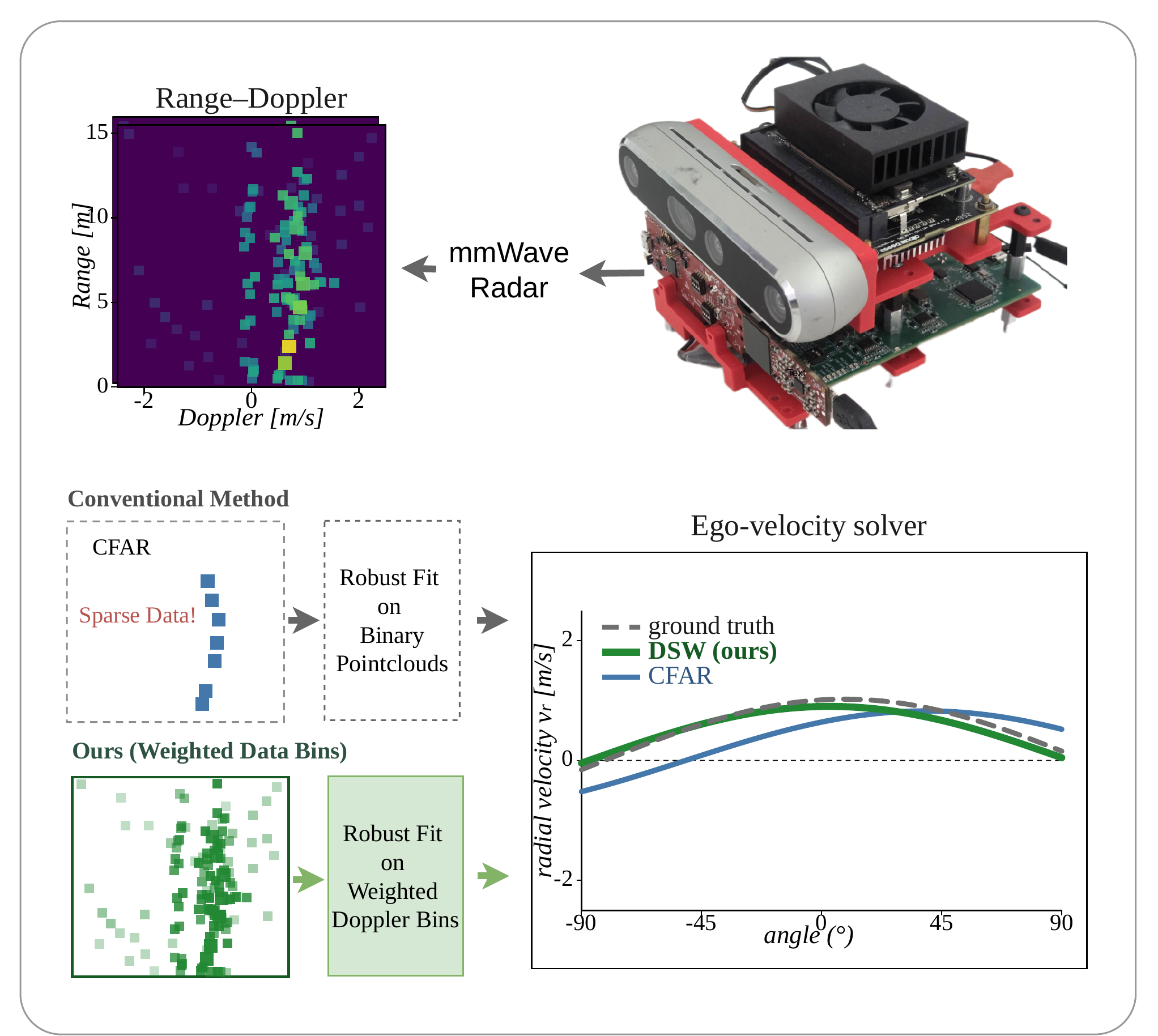}
    \caption{Motivating example for the dense radar ego-velocity front end.
CFAR reduces a range-Doppler frame to sparse binary detections before fitting
ego-velocity. The proposed system instead fits ego-velocity from dense weighted Doppler
evidence, producing a robust fit closer to the ground-truth velocity relation in
this representative frame.}
    \label{fig:hero}
    \Description{A single mmWave radar frame is shown as a range-Doppler map.
The conventional front end applies CFAR and fits ego-velocity from sparse binary
detections. The proposed front end keeps dense weighted Doppler bins, and its solver
curve lies closer to the ground-truth velocity curve.}
\end{figure}

\section{Introduction}
\label{sec:introduction}

Millimeter-wave frequency-modulated continuous-wave radar sensing offers valuable
range, angle, and radial-velocity information for robots in darkness, fog, dust,
and smoke~\cite{Harlow2024Survey}. Therefore, radar-inertial odometry can
support robust motion estimation when visual sensing becomes unreliable.
For single-chip radar, ego-velocity estimation must turn dense
range-Doppler-angle spectra into trustworthy velocity and uncertainty
measurements. Existing methods usually handle this either by first reducing
the spectrum to sparse detections, or by learning dense weights, motion, or
uncertainty from training data matched to the sensing platform.

Conventional pipelines apply constant false alarm rate (CFAR) detection, an
adaptive thresholding rule whose output is a binary keep/discard decision, and
then estimate ego-velocity with least squares or random sample consensus
(RANSAC)~\cite{Kellner2013, Doer2020rio, Doer2021xRIO, Kim2025EKFRIOTC}.
While this design is efficient and established, the binary output removes
non-forward facing weak cells before estimation (\cref{fig:hero}). Sparse refinements sharpen
angle estimates, upsample detections, reweight point clouds, or add inertial
constraints~\cite{Jiang2025DBE, RadarHD2023, RadarSFD2026, CREVE2024,
VGCRIO2025}. These methods improve the surviving detections. However, they still
depend on a detector to decide which spectral cells reach the estimator. Only
recently, learning-based dense systems replace the detector with learning-based
weights or motion regression~\cite{milliEgo2020, BatMobility2023,
Radarize2024, UNRIO2026}, reaching strong accuracy when trained on data
matched to the radar chip, antenna layout, and environment. In contrast,
Doppler-aware radar odometry aligns whole scans analytically. However, it targets
scan-to-local-map registration on a spinning radar~\cite{Gentil2025DRO}, a
different hardware family from single-chip mmWave.

These trade-offs expose three practical challenges for deployable
single-chip radar-inertial odometry. First, the front-end should retain weak
off-boresight Doppler evidence because these cells improve velocity geometry
and can reduce drift after inertial fusion. Second, it should report a usable
velocity covariance so the measurement can plug into a standard inertial
back-end, rather than requiring a private pose regressor. Third, it should
remain lightweight enough for embedded platforms by avoiding per-platform
training and heavy network inference. The research problem is therefore to
assign per-cell confidence, robust ego-velocity, and covariance directly from
dense range-Doppler-angle spectra without collapsing them to sparse detections.

In this work, we introduce \emph{Dense Soft Weighting} (DSW), an
analytic radar front-end for direct ego-velocity estimation from dense
range-Doppler spectra.
\begin{itemize}
    \item We introduce a dense per-cell confidence field for radar
    ego-velocity estimation. The confidence is derived from beam-domain power
    and concentration, turning each range-Doppler cell into a weighted motion
    cue instead of a binary detector output. This keeps sub-threshold Doppler
    evidence available while reducing confidence in ambiguous cells, producing
    a dense measurement population rather than a detector-selected point
    cloud.
    \item We produce a robust ego-velocity measurement with covariance for a
    standard inertial back-end. Robust weighted least squares lets weak returns
    contribute at reduced confidence while suppressing residual outliers. The
    covariance exposes residual, geometric, and Doppler-bin uncertainty to the
    filter.
    \item We show that the same dense front-end transfers across radar
    configurations and antenna layouts. The evaluation spans two public
    datasets and our sensor rig, compares against point-cloud RIO and
    learning-based odometry baselines, and keeps a shared inertial back-end for
    the analytic methods so front-end effects remain isolated.
\end{itemize}

The remainder of this paper is organized as follows.
\Cref{sec:related-work} positions DSW against sparse point-cloud and learned
dense radar front ends. \Cref{sec:method} derives the dense weighting, robust
ego-velocity solver, and covariance, while \Cref{sec:experiments,sec:results}
describe the evaluation protocol and report accuracy, ablation, and runtime
results. \Cref{sec:conclusion} concludes with limitations and deployment
directions.

\section{Related Work}
\label{sec:related-work}

\subsection{Sparse Point-Cloud Ego-Velocity Estimation}
\label{sec:rw-sparse}

Sparse radar ego-velocity front-ends share a common interface: the radar
spectrum is first reduced to a small set of detections, and motion is
then estimated from the Doppler values attached to those points. This
detector-first design is widely used because it gives the estimator a
compact geometric measurement set and fits naturally into robust
least-squares pipelines. Kellner et al.~\cite{Kellner2013} established
the single-scan formulation by relating radial velocity measurements to
three-dimensional ego-velocity and solving the resulting linear system on
automotive radar detections. Across Doer et al.'s radar-odometry
work~\cite{Doer2020rio, Doer2020calib, Doer2021yaw, Doer2021xRIO}, the
front end remains a sparse point-cloud velocity estimator: three
detections generate a RANSAC hypothesis, least squares refines the inlier
set, and the velocity covariance is fitted from the inlier residuals.
CREVE~\cite{CREVE2024} and VGC-RIO~\cite{VGCRIO2025} retain this
point-cloud interface while changing the robustification mechanism, using
IMU-derived acceleration constraints and spatial density-based weighting,
respectively. MRIO uses a robust Cauchy cost optimized by
Levenberg--Marquardt in place of RANSAC sampling~\cite{Huang2024MRIO},
while RIV-SLAM adds multi-strategy weighted least squares and
azimuth-dependent weighting~\cite{RIVSLAM2024}.

These methods improve the estimator, the robust loss, or the fusion
around the detected points, but the measurement population is still fixed
by CFAR or a related detector before velocity estimation begins.
Sub-threshold cells, weak off-boresight returns, and useful sidelobe
structure have already been discarded, and covariance must be calibrated
from a comparatively small residual set. The limitation is also
practical. In single-chip systems the point cloud is typically generated
by the radar's onboard DSP rather than by host-side processing of the
full ADC cube; compute, memory, and bandwidth budgets favor simplified
CFAR and angle estimation, reduced-precision intermediates, and quantised
output. The range-Doppler-angle spectrum is thus pruned or compressed
before it reaches the estimator, and CFAR tuning becomes a sensitive,
scene-dependent trade-off between admitting clutter and discarding weak
but geometrically useful cells. The estimator inherits both the
detector's sparsity and the signal processor's implementation choices.

\begin{figure*}[t]
    \centering
    \includegraphics[width=\textwidth]{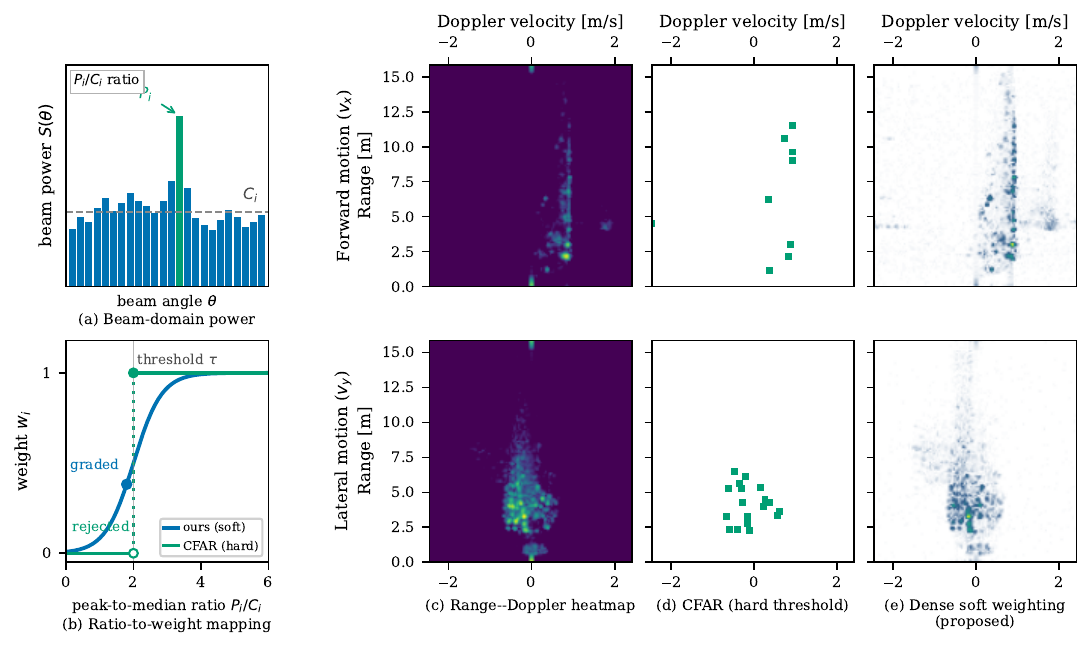}
    \caption{Soft weighting versus binary thresholding. Left: (a)~beam-domain
    power at one $(r,d)$ range-Doppler cell and (b)~the mapping from
    peak-to-median ratio to cell weight. Right: real ColoRadar frames
    (\texttt{arpg\_lab\_run0}, forward-dominant top and lateral-dominant
    bottom; Doppler velocity on the horizontal axis, range on the vertical
    axis), showing (c)~range-Doppler power, (d)~CASO-CFAR binary detections used
    by the point-cloud baseline, and (e)~a continuous per-cell soft weight
    normalised to the per-frame maximum.}
    \label{fig:weighting}
    \Description{Left: a beam-domain power spectrum at one range-Doppler cell
    with its peak and cross-beam median, and the sigmoid mapping from the
    peak-to-median ratio to a graded weight, contrasted with a binary CFAR threshold. Right: two
    real ColoRadar frames showing range-Doppler power, the sparse CASO-CFAR
    detections a binary detector keeps, and a continuous per-cell soft weight,
    which grades roughly an order of magnitude more cells than the detector keeps.}
\end{figure*}

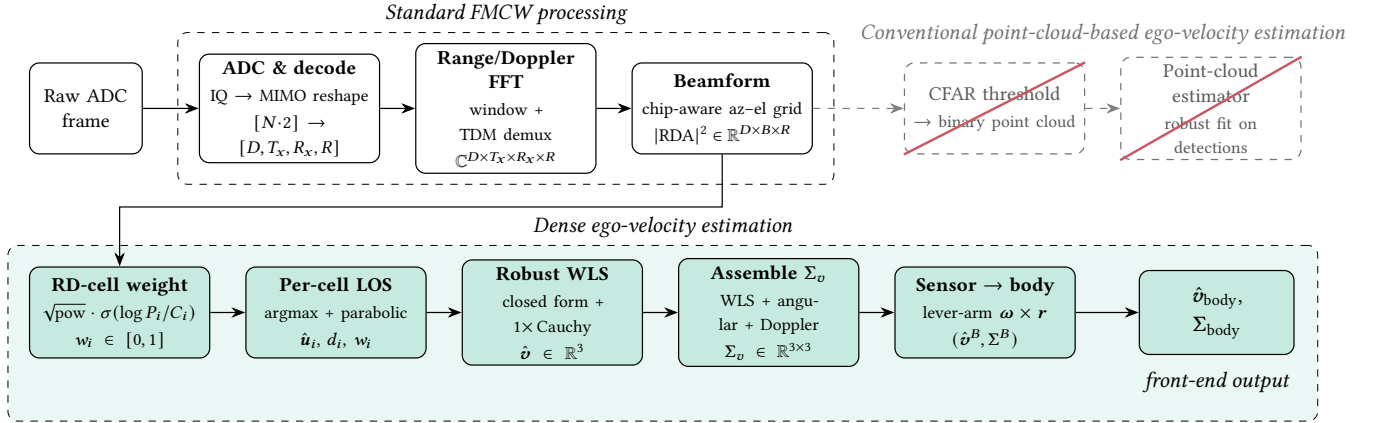
\begin{figure*}[t]
  \centering
  \resizebox{\textwidth}{!}{%
  \begin{tikzpicture}[
      font=\small,
      >={Stealth[length=2mm]},
      line width=0.5pt,
      src/.style={draw,rounded corners,align=center,inner sep=3pt,
                  minimum height=13mm,text width=14mm},
      m/.style={draw,rounded corners,align=center,inner sep=2.5pt,
                minimum height=13mm,text width=24mm},
      mine/.style={draw,rounded corners,align=center,inner sep=2.5pt,
                minimum height=13mm,text width=24mm,fill=OIGreen!22},
      pill/.style={draw,rounded corners,align=center,inner sep=3pt,
                   minimum height=12mm,text width=20mm,fill=OIGreen!22},
      ghost/.style={draw=black!55,dashed,rounded corners,align=center,
                    inner sep=2.5pt,minimum height=13mm,text width=24mm,text=black!65},
    ]
    \node[src] (raw) {Raw ADC\\frame};
    \node[m, right=8mm of raw] (s1)
          {\textbf{ADC \& decode}\\[1pt]{\footnotesize IQ $\to$ MIMO reshape}\\[1pt]{\footnotesize $[N{\cdot}2]\!\to\![D,T_x,R_x,R]$}};
    \node[m, right=5mm of s1] (s2)
          {\textbf{Range/Doppler FFT}\\[1pt]{\footnotesize window + TDM demux}\\[1pt]{\footnotesize $\mathbb{C}^{D\times T_x\times R_x\times R}$}};
    \node[m, right=5mm of s2] (s3)
          {\textbf{Beamform}\\[1pt]{\footnotesize chip-aware az--el grid}\\[1pt]{\footnotesize $|\mathrm{RDA}|^2\!\in\!\mathbb{R}^{D\times B\times R}$}};
    \node[ghost, right=13mm of s3] (cfar)
          {CFAR threshold\\[1pt]{\footnotesize $\to$ binary point cloud}};
    \node[ghost, right=5mm of cfar] (ransac)
          {Point-cloud estimator\\[1pt]{\footnotesize robust fit on detections}};
    \node[font=\normalsize\itshape, text=black!55, align=center]
          at ($(cfar.north)!0.5!(ransac.north)+(0,3.6mm)$) (cfarlbl)
          {Conventional point-cloud-based ego-velocity estimation};
    \node[mine, anchor=north west] (s4) at ([yshift=-16mm]raw.south west)
          {\textbf{RD-cell weight}\\[1pt]{\footnotesize $\sqrt{\smash[b]{\text{pow}}}\cdot\sigma(\log P_i/C_i)$}\\[1pt]{\footnotesize $w_i\in[0,1]$}};
    \node[mine, right=5mm of s4] (s5)
          {\textbf{Per-cell LOS}\\[1pt]{\footnotesize argmax + parabolic}\\[1pt]{\footnotesize $\hat{\bm u}_i,\,d_i,\,w_i$}};
    \node[mine, right=5mm of s5] (s6)
          {\textbf{Robust WLS}\\[1pt]{\footnotesize closed form + 1$\times$\,Cauchy}\\[1pt]{\footnotesize $\hat{\bm v}\in\mathbb{R}^3$}};
    \node[mine, right=5mm of s6] (s7)
          {\textbf{Assemble $\bm\Sigma_v$}\\[1pt]{\footnotesize WLS + angular + Doppler}\\[1pt]{\footnotesize $\bm\Sigma_v\in\mathbb{R}^{3\times3}$}};
    \node[mine, right=5mm of s7] (s8)
          {\textbf{Sensor $\to$ body}\\[1pt]{\footnotesize lever-arm $\bm\omega\times\bm r$}\\[1pt]{\footnotesize $(\hat{\bm v}^B,\bm\Sigma^B)$}};
    \node[pill, font=\normalsize, right=9mm of s8] (out) {$\hat{\bm v}_{\text{body}}$,\\[1pt]$\bm\Sigma_{\text{body}}$};
    \node[font=\normalsize\itshape] (outlbl) [below=1.2mm of out] {front-end output};
    \draw[->] (raw) -- (s1);
    \draw[->] (s1) -- (s2);
    \draw[->] (s2) -- (s3);
    \draw[->] (s3.south) -- ++(0,-7.5mm) -| (s4.north);          %
    \draw[->, dashed, black!50] (s3.east) -- (cfar.west);         %
    \draw[->, dashed, black!50] (cfar.east) -- (ransac.west);
    \draw[TolBrightRed!85!black, line width=1.0pt] (cfar.south west) -- (cfar.north east); %
    \draw[TolBrightRed!85!black, line width=1.0pt] (ransac.south west) -- (ransac.north east); %
    \draw[->] (s4) -- (s5);
    \draw[->] (s5) -- (s6);
    \draw[->] (s6) -- (s7);
    \draw[->] (s7) -- (s8);
    \draw[->] (s8) -- (out);
    \begin{scope}[on background layer]
      \node[draw, dashed, rounded corners, fit=(s1)(s3), inner sep=3mm,
            label={[font=\normalsize\itshape]above:Standard FMCW processing}] {};
      \node[draw, dashed, rounded corners, fit=(s4)(s8)(out)(outlbl), inner sep=3mm, fill=OIGreen!8,
            label={[font=\normalsize\itshape]above:Dense ego-velocity estimation}] {};
    \end{scope}
  \end{tikzpicture}%
  }
  \caption{System overview of a dense radar front-end. A standard FMCW
    chain produces a dense range-Doppler-angle representation. The greyed
    branch shows the conventional CFAR-to-point-cloud route; the dense
    branch keeps the dense representation, assigns per-cell weights,
    constructs line-of-sight vectors, estimates velocity with robust WLS,
    assembles a covariance, and expresses the result in the body frame.
    The front-end emits only
    $(\hat{\bm v}_{\text{body}}, \bm\Sigma_{\text{body}})$.}
  \label{fig:system-overview}
  \Description{Block diagram of a dense radar front-end. A standard FMCW
    chain (ADC decode, range/Doppler FFT, beamforming) forks: a greyed,
    crossed-out branch marks the conventional CFAR detection step that the dense front-end
    bypasses, while the main branch feeds the dense estimator
    (cell weighting, line-of-sight construction, robust weighted least
    squares, covariance assembly, sensor-to-body lever-arm), ending at the
    body-frame velocity and covariance output. No back-end filter is shown.}
\end{figure*}

\subsection{Dense and Direct Radar Front-Ends}
\label{sec:rw-direct}

A complementary direction is to reduce the information loss introduced by
detector-first pipelines by operating closer to the radar spectrum. Existing
analytic approaches follow two main paths: improving the signal processing
before point-cloud extraction, or formulating odometry directly on radar
intensity.

Digital-beamforming-enhanced radar odometry~\cite{Jiang2025DBE} follows the
first path, using raw ADC captures and host-side digital beamforming to produce
higher-quality point clouds for ego-velocity estimation. However, DBE retains
the point-cloud interface: improved signal processing raises detection quality,
but the detector still determines which spectral cells reach the estimator. DRO~\cite{Gentil2025DRO} follows the second path by aligning dense radar
intensity scans without learned components. It is designed for mechanically
scanning automotive radar and performs Doppler-aware \mbox{SE(2)}
scan-to-local-map registration on dense $360^\circ$ scans. This formulation is
effective for spinning radar, but it does not directly transfer to compact
single-chip mmWave phased arrays, which provide range-Doppler-angle spectra
rather than planar panoramic scans. Dense single-chip ego-velocity estimation
therefore presents a different interface problem: spectral evidence must be
converted into velocity weights and covariance without first reducing the radar
cube to detector-selected points.

\subsection{Learning-Based Radar Ego Motion Approaches}
\label{sec:rw-dense}

Recent work on single-chip radar ego-motion has increasingly used
learning to predict motion, weights, or uncertainty directly from radar
data. milliEgo~\cite{milliEgo2020} studies learning-based single-chip
radar-inertial ego-motion, fusing mmWave radar and inertial measurements
in an end-to-end network that regresses ego-motion. Its input is the
detected point cloud rather than the dense spectrum, so the network
operates on the same sparse interface as conventional pipelines. Learning has also been applied at the
measurement level: RadarHD~\cite{RadarHD2023} trains a network with
lidar ground truth to convert low-resolution single-chip radar heatmaps
into lidar-like high-resolution point clouds, improving the detections
that downstream estimators consume rather than the motion estimate
itself.
BatMobility~\cite{BatMobility2023} instead estimates ego-velocity from
the dense range-Doppler spectrum, learning a radar-flow representation
from Doppler-angle data without intermediate detection.
Radarize~\cite{Radarize2024} extends this line to the full spectrum,
training separate translation and rotation networks on the same dense
range-Doppler input to produce odometry. 4DEgo~\cite{4DEgo2023} and
S$^3$E~\cite{S3E2025} likewise learn from radar heatmaps or spectra
fused with inertial data, while UNRIO~\cite{UNRIO2026} is closest in
estimator structure: it predicts a Doppler image with a transformer,
estimates velocity with weighted least squares, and learns an
uncertainty head for the downstream graph. Together, these methods
show that pre-detection radar data carries useful motion information.

Despite this progress, several deployment issues remain open for
learning-based front-ends. Models trained for a particular radar chip and
chirp configuration may not transfer directly to other sensors, since the
learned representation depends on the antenna geometry and
range-Doppler resolution of the training platform. Performance can also
depend on the training environment, as clutter and multipath statistics
vary across scenes. In addition, real-time operation on the embedded
platforms typically paired with single-chip radar remains comparatively
unexplored. These considerations distinguish learning-based dense
front-ends from analytic formulations in which per-cell weights and
covariance are derived from the radar spectrum itself.

\subsection{Radar-Inertial Odometry Backends}
\label{sec:rw-backends}

Once a radar front-end produces an ego-velocity measurement, radar-inertial
odometry becomes a fusion problem: the backend must combine that velocity
with inertial propagation, calibration states, and temporal alignment. A
substantial body of RIO work therefore focuses on the downstream estimator.
The EKF-RIO family places a sparse radar velocity front-end inside an ESKF
and extends the filter with online extrinsic calibration~\cite{Doer2020rio,
Doer2020calib}, Manhattan-world yaw aiding~\cite{Doer2021yaw}, and
multi-radar fusion~\cite{Doer2021xRIO}. Other backends refine the state
and coupling: Kim et al.'s online temporal-calibration
formulation~\cite{Kim2025EKFRIOTC} and temporal-delay analysis for
RIO~\cite{Stironja2025temporal} estimate the radar-to-IMU temporal
offset $t_d$, tightly-coupled EKF-RIO~\cite{Michalczyk2022tight}
updates the filter with individual radar detections rather than a single
per-scan velocity, while Yang et al.~\cite{GoRIO2025} and Huang et
al.~\cite{Huang2024less} add ground-plane and IMU-aided velocity priors.

These methods advance the fusion and filtering stage, but they generally
treat the radar ego-velocity input as an upstream measurement supplied by
a front-end. The resulting odometry is therefore still constrained by the
quality and covariance calibration of that input, independently of the
backend formulation.

\section{Method}
\label{sec:method}

\subsection{System Overview}
\label{sec:overview}

The proposed system maps a raw single-chip radar frame and IMU stream to an
inertially fused trajectory. \Cref{fig:system-overview} summarises this
data flow: the radar branch decodes an ADC frame into MIMO channels,
applies range and Doppler FFTs with TDM demultiplexing, beamforms onto a
chip-aware azimuth--elevation grid, and reduces the result to a dense
power tensor. DSW assigns each range-Doppler cell a continuous confidence,
refines its line-of-sight (LOS) vector, estimates sensor-frame velocity
with robust WLS, assembles a covariance, and maps the result to the body
frame using the calibrated sensor-to-body transform and the gyro-driven
lever-arm term $\bm\omega\times\bm r$. The front-end emits only
$(\hat{\bm v}^{B},\bm\Sigma^{B})$; the ESKF fuses this body-frame
measurement with IMU propagation to produce the evaluated odometry
trajectory. Unlike sparse Doppler pipelines that detect a point cloud and
reject outliers with three-point
RANSAC~\cite{Kellner2013, Doer2020rio, Doer2021xRIO, Kim2025EKFRIOTC},
DSW retains every range-Doppler cell at a continuous weight and resolves
outliers with a single analytic reweighting; we examine the implications
of this choice in \cref{sec:discussion}.

\subsection{Radar Signal Processing Front-End}
\label{sec:preliminaries}

We decode each FMCW frame from IQ samples and reshape it into $D$ chirps,
$T_x$ transmitters, $R_x$ receivers, and $R$ range samples. Applying a
range FFT along fast time and a Doppler FFT along slow time, together with
TDM demultiplexing, yields
$\mathbb{C}^{D\times T_x\times R_x\times R}$. We then beamform over the
virtual array onto a two-dimensional grid of $B=B_\alpha B_\beta$ beams,
using the chip's TX/RX layout rather than a generic transform; $\alpha$
and $\beta$ denote azimuth and elevation. DSW consumes the dense power
tensor
\begin{equation}
    \mathcal{P}(d,b,r) = |\mathrm{RDA}(d,b,r)|^2
    \in \mathbb{R}^{D\times B\times R},
    \label{eq:rad-tensor}
\end{equation}
where $d$ indexes Doppler bins, $b=(b_\alpha,b_\beta)$ indexes the
azimuth--elevation beam, and $r$ indexes range. From
\cref{eq:rad-tensor}, each range-Doppler cell $i=(d_i,r_i)$ supplies a
beam-domain power spectrum over $b$, from which DSW extracts the Doppler
velocity $d_i$, the LOS vector $\uvec{u}_i$, and the weight $w_i$.

\subsection{Radar Ego-Velocity Measurement Model}
\label{sec:meas-model}

Each range-Doppler cell contributes one LOS Doppler constraint. For a
cell $i$, let $\dopp_i$ be its Doppler radial velocity and
$\uvec{u}_i\in\mathbb{R}^3$ be the unit LOS vector in the radar sensor
frame. We model
\begin{equation}
    \dopp_i \;=\; -\uvec{u}_i^\transp \egovel + \epsilon_i,
    \qquad\text{(per cell $i$)}
    \label{eq:doppler-model}
\end{equation}
where $\egovel\in\mathbb{R}^3$ is the sensor-frame ego-velocity and
$\epsilon_i$ aggregates thermal noise, clutter, sidelobes, and the radial
velocity of any moving reflector contributing to the cell. Because a
single cell constrains $\egovel$ only along $\uvec{u}_i$, we estimate
three-dimensional velocity from many weighted LOS directions across the
azimuth--elevation grid. Concatenating the per-cell constraints over all
$N$ cells yields the stacked system
\begin{equation}
    \vect{d} \;=\; -\mat{U}\,\egovel + \boldsymbol{\epsilon},
    \qquad\text{(stacking $N$ dense cells)}
    \label{eq:stacked-meas}
\end{equation}
where $\mat{U}=[\uvec{u}_1,\,\ldots,\,\uvec{u}_N]^\transp\in\mathbb{R}^{N\times3}$
stacks the per-cell LOS directions and $\vect{d},\boldsymbol{\epsilon}\in\mathbb{R}^N$
stack the corresponding Doppler velocities and noise terms, while the
shared unknown $\egovel\in\mathbb{R}^3$ couples every cell.

\subsection{Range-Doppler Cell Confidence Weight}
\label{sec:dense-estimator}

A detector commits to a binary keep/discard decision before the velocity
geometry is known. We instead score each cell by the two quantities a
CFAR detector already inspects---how much power the cell carries and how
tightly it concentrates in one beam---and map them to a continuous
weight.

For cell $i=(d_i,r_i)$, the beam-domain power spectrum is
$S_i(b)=\mathcal{P}(d_i,b,r_i)$. We take its peak $P_i=\max_b S_i(b)$ and
its cross-beam median $C_i=\med_b S_i(b)$ and form
{\setlength\abovedisplayskip{3pt plus 1pt minus 1pt}%
 \setlength\belowdisplayskip{3pt plus 1pt minus 1pt}%
\begin{align}
    w_i &= u_i v_i,
    \label{eq:weight}
\end{align}}
where $w_i$ is the confidence assigned to the range-Doppler cell. The
first factor is the normalized power score
{\setlength\abovedisplayskip{3pt plus 1pt minus 1pt}%
 \setlength\belowdisplayskip{3pt plus 1pt minus 1pt}%
\begin{align}
    u_i &= \sqrt{P_i / \max_j P_j},
    \label{eq:weight-power}
\end{align}}
where the maximum $\max_j P_j$ runs over all range-Doppler cells in the
current frame. The score $u_i\in[0,1]$ makes the weighting
scale-invariant and, through the square root, limits the leverage of a
single bright reflector. The second factor is the concentration gate
{\setlength\abovedisplayskip{3pt plus 1pt minus 1pt}%
 \setlength\belowdisplayskip{3pt plus 1pt minus 1pt}%
\begin{align}
    v_i &= \sigma\!\left(\frac{\log(P_i/C_i) - \log\tau}{\kappa}\right),
    \label{eq:weight-concentration}
\end{align}}
where $\sigma(x)=1/(1+e^{-x})$. The gate $v_i\in[0,1]$ is a sigmoid of
the log peak-to-median ratio: $\tau$ is the ratio at which $v_i=1/2$ and
$\kappa$ sets the slope. We fix $\tau=200$ and $\kappa=0.5$ on every
dataset, with no per-platform tuning.

The ratio $P_i/C_i$ is the cross-beam analogue of a cell-averaging CFAR
test statistic, with the median $C_i$ standing in for the local
noise-floor estimate; the sigmoid turns the detector's hard threshold
into a soft gate centred at $\tau$. A cell whose energy concentrates in
one direction has $P_i \gg C_i$ and $v_i \to 1$, whereas a sidelobe- or
clutter-dominated cell spreads its energy across beams, giving
$P_i \sim C_i$ and $v_i \to 0$. Cells below a detector's threshold can
thus still contribute at reduced weight when they improve the velocity
geometry (\cref{fig:weighting}(a,b)), which on real frames yields a dense
weighted measurement set rather than the sparse CFAR detections of
\cref{fig:weighting}(d,e). Because the peak-to-median ratio rarely
exceeds ${\sim}100\times$, below $\tau=200$, a hard threshold at $\tau$
would retain too few cells and can make the dense estimate degenerate
(quantified by the ablation in \cref{sec:ablation}, \cref{tab:ablation}).

\subsection{Per-Cell Line of Sight and Robust Velocity Estimation}
\label{sec:wls}

Each weighted cell contributes one LOS Doppler constraint, but the beam
grid quantises its direction, most severely along the axis with the
smallest aperture and hence the coarsest angular resolution; on
azimuth-dominant arrays this is elevation. We therefore refine each peak
to sub-bin resolution before forming $\uvec{u}_i$. We locate the dominant beam
\begin{equation}
    \hat{b}_i=(\hat{b}_{\alpha,i},\hat{b}_{\beta,i})=\arg\max_b S_i(b),
    \label{eq:argmax-beam}
\end{equation}
and refine it independently along the azimuth and elevation axes of the
beam grid by standard three-point parabolic interpolation on the peak bin
power and its two neighbours. Refinement matters most along whichever axis the array resolves most
coarsely---elevation on azimuth-dominant chips, but equally any poorly
resolved direction on an antenna array with a sparse aperture---where
bin-quantised angles would otherwise leave the corresponding velocity
component poorly observed. Converting the refined angles
from spherical to Cartesian coordinates gives the unit line-of-sight
vector $\uvec{u}_i$.

With per-cell directions and confidences in hand, we stack the rows of
\cref{eq:stacked-meas}, weight each by its cell confidence, and solve the
resulting weighted least-squares problem in closed form,
\begin{equation}
    \hat{\egovel}^{(0)}
    \,=\, -\bigl(\mat{U}^\transp \mat{W}\,\mat{U}\bigr)\inv
        \mat{U}^\transp \mat{W}\,\vect{d},
    \qquad
    \mat{W} = \mathrm{diag}(w_1,\ldots,w_N).
    \label{eq:wls}
\end{equation}
The confidence weights already suppress sidelobe-dominated cells, but
moving reflectors and multipath can still inject residual outliers. To
reject them, we run a single iteratively-reweighted-least-squares pass
with a Cauchy loss. From the residuals
\begin{equation}
    r_i = \dopp_i + \uvec{u}_i^\transp \hat{\egovel}^{(0)},
    \label{eq:residual}
\end{equation}
we estimate a robust scale $\hat{\sigma}_r$ from the median absolute
deviation of $\vect{r}$ and form the Cauchy weights
\begin{equation}
    \rho_i = \frac{1}{1 + (r_i / c\,\hat{\sigma}_r)^2},
    \label{eq:cauchy}
\end{equation}
collected in $\boldsymbol{\rho} = \mathrm{diag}(\rho_1,\ldots,\rho_N)$. We
then re-evaluate \cref{eq:wls} with $\mat{W}\boldsymbol{\rho}$ in place of
$\mat{W}$. We use a single fixed cutoff $c = 2$ for both apertures, and
we stop after one pass because further iterations changed per-frame RMSE
by less than $10^{-4}$~\si{\meter\per\second} in our pilot studies.

\subsection{Velocity Covariance Assembly}
\label{sec:fisher}

The proposed system reports the sensor-frame covariance $\bm\Sigma_v^{R}$ alongside each
velocity estimate. We follow sparse Doppler ego-velocity
estimators~\cite{Kellner2013, Doer2020rio} in reporting the closed-form
least-squares covariance of the velocity fit, and extend it with
beam-direction uncertainty and Doppler-bin quantisation:
\begin{equation}
    \bm\Sigma_v^{R} \,=\, \bm\Sigma_v^{\mathrm{wls}}
    \;+\; \bm\Sigma_v^{\mathrm{ang}} \;+\; \bm\Sigma_v^{\mathrm{dop}},
    \label{eq:cov-v}
\end{equation}
with the terms
\begin{align}
    \bm\Sigma_v^{\mathrm{wls}} &= \hat{\sigma}_r^2\,\bigl(\mat{U}^\transp \mat{W}\boldsymbol{\rho}\,\mat{U}\bigr)\inv, \label{eq:cov-wls}\\
    \bm\Sigma_v^{\mathrm{ang}} &= \|\hat{\egovel}\|^2\,\mat{J}\,\mathrm{diag}(\sigma_\alpha^2,\sigma_\beta^2)\,\mat{J}^\transp, \label{eq:cov-ang}\\
    \bm\Sigma_v^{\mathrm{dop}} &= \sigma_{\mathrm{dop}}^2\,\Identity. \label{eq:cov-dop}
\end{align}
The three terms account, respectively, for the uncertainty of the
velocity fit (the residual variance $\hat{\sigma}_r^2$ carried through the
least-squares geometry), the angular uncertainty of the beam directions
(the per-axis pointing deviations $\sigma_\alpha, \sigma_\beta$ mapped
through the line-of-sight Jacobian $\mat{J}$ and growing with speed
$\|\hat{\egovel}\|$), and a floor $\sigma_{\mathrm{dop}}^2$ set by the
radar's finite Doppler resolution, the quantisation noise of reading the
radial velocity to the nearest Doppler bin of width
$\Delta v_{\mathrm{bin}}$. All three are computed in closed form from the
radar configuration and the measured cells, with no per-platform tuning,
so the covariance transfers across radars; the vertical term is enlarged
on arrays with few elevation rows, where $v_z$ is weakly observed.

The resulting estimate is first obtained in the radar sensor frame.
Before fusion, the front-end maps both the velocity mean and covariance
to the body frame shown in \cref{fig:system-overview}. For a radar
mounted at a fixed lever arm $\bm r$ from the body origin, platform
rotation induces the sensor-frame apparent velocity
$\bm\omega\times\bm r$. We remove this term using the measured gyroscope
rate and then apply the calibrated sensor-to-body rotation. This
conversion is performed once, at the front-end output, so the ESKF
receives $(\hat{\bm v}^{B},\bm\Sigma^{B})$ directly and applies no
additional lever-arm correction.

\subsection{IMU Fusion Back-End}
\label{sec:eskf}

To fuse each radar velocity measurement $(\hat{\bm v}^{B},\bm\Sigma^{B})$
with inertial propagation and recover the full trajectory, we use a
loose-coupled ESKF. The filter, its process noise, and its innovation
gate are identical across every analytic method evaluated, so
trajectory-level differences are attributable to the front-end. We
follow the loose-coupled radar-velocity formulation of
EKF-RIO~\cite{Doer2020rio} and add online radar-to-IMU temporal
calibration following Kim et al.~\cite{Kim2025EKFRIOTC}. We parameterise
the attitude error in the quaternion-error form of
Sol\`a~\cite{Sola2017quaternion}. The nominal state
\begin{equation}
    \state \,=\, \bigl[\,
    \vect{p}^\transp,\;
    \vect{v}^\transp,\;
    \vect{q}^\transp,\;
    \vect{b}_a^\transp,\;
    \vect{b}_g^\transp,\;
    t_d \bigr]^\transp
    \label{eq:nominal-state}
\end{equation}
holds position, velocity, attitude quaternion, accelerometer and
gyroscope biases, and a radar-to-IMU temporal offset $t_d$ that we
estimate online following Kim et al.~\cite{Kim2025EKFRIOTC}; because the
quaternion error lives in the 3-DoF tangent space, the propagated error
state $\errorstate$ is 16-dimensional. We calibrate the
radar-to-IMU spatial extrinsic offline;
the proposed front-end consumes its lever-arm velocity correction before the
measurement reaches the filter, as described in \cref{sec:fisher}. We
propagate the error state with inertial measurements through the
standard discrete-time dynamics
\begin{align}
    \errorstate_{k|k-1} &\,=\, \Fmat_k\,\errorstate_{k-1|k-1}, \notag \\
    \cov_{k|k-1} &\,=\, \Fmat_k\,\cov_{k-1|k-1}\,\Fmat_k^\transp
    + \Gmat_k\,\Qmat\,\Gmat_k^\transp,
    \label{eq:eskf-prediction}
\end{align}
and at each radar epoch we apply the body-frame velocity estimate as a
partial-state measurement update,
\begin{equation}
    \meas_k \,=\, \hat{\bm v}^{B}_k \,=\, \Hmat\,\errorstate_k + \noise_k,
    \qquad \noise_k \sim \mathcal{N}(\vect{0},\,\bm\Sigma^{B}_k),
    \label{eq:eskf-update}
\end{equation}
where $\Hmat$ is the velocity-measurement Jacobian for the body-frame
velocity implied by the inertial state; it applies no second lever-arm
correction. We gate each update with a Mahalanobis $\chi^2_3$ test at
the $0.95$ level, rejecting innovations inconsistent with
$\bm\Sigma^{B}_k$. We adopt the IMU process-noise parameterisation of
Kim et al.~\cite{Kim2025EKFRIOTC}, so every analytic method evaluated in
this paper shares an identical $\Qmat$ specification and the comparison
reduces to the choice of front-end.

\section{Evaluation}
\label{sec:experiments}

\begin{table}[t!]
    \centering
    \caption{Datasets and radar chirp configuration used for evaluation.}
    \label{tab:datasets}
    \renewcommand{\arraystretch}{1.0}
    \setlength{\tabcolsep}{4pt}
    \footnotesize
    \begin{tabular}{@{}
        >{\raggedright\arraybackslash}p{2.75cm}
        >{\centering\arraybackslash}p{1.4cm}
        >{\centering\arraybackslash}p{1.4cm}
        >{\centering\arraybackslash}p{1.4cm}@{}}
    \toprule
     & \multicolumn{3}{c}{\textbf{Dataset}} \\
    \cmidrule(l){2-4}
    \textbf{Property} &
    \makecell{\textbf{ColoRadar}\\\cite{Kramer2022ColoRadar}} &
    \makecell{\textbf{Radarize}\\\cite{Radarize2024}} &
    \textbf{Self-collected Dataset} \\
    \midrule
    Platform        & Handheld        & Handheld, cart, robot & Handheld \\
    Radar chip      & AWR1843 BOOST   & AWR1843 BOOST         & \shortstack{AWR6843\\AOP-EVM} \\
    Data capture    & DCA1000         & DCA1000               & DCA1000 \\
    IMU             & 3DM-GX5-25      & Bosch BMI055          & Xsens MTi-320 \\
    Ground truth    & LiDAR, MoCap    & VIO                   & MoCap \\
    Environment     & Indoor, outdoor & Indoor                & Indoor \\
    Motion & 2D         & 2D       & \textbf{3D} \\
    Center freq.\ $f_c$ [GHz]                         & 77.6        & 78.7        & 61.0 \\
    Bandwidth $B$ [GHz]                               & 1.20        & 3.36        & 1.91 \\
    ADC rate $f_s$ [MHz]                              & 10.67       & 2.29        & 4.68 \\
    Ramp $T_\mathrm{ramp}$ [\si{\micro\second}]       & 20.0        & 50.0        & 63.75 \\
    Idle $T_\mathrm{idle}$ [\si{\micro\second}]       & 110         & 122         & 38.44 \\
    Chirp rep.\ $T_\mathrm{PRT}$ [\si{\micro\second}] & 130         & 172         & 102.19 \\
    Frame $T_f$ [ms]                                  & 100         & 33.3        & 100 \\
    Samples/chirp $N_s$                               & 128         & 96          & 255 \\
    Chirps/frame $N_c$                                & 128         & 32          & 60 \\
    Antenna $N_\mathrm{tx}/N_\mathrm{rx}$             & $3\times4$  & $3\times4$  & $3\times4$ \\
    Az./El. FoV [$^\circ$]                                        & 60/15       & 60/15       & 60/60 \\
    \bottomrule
    \end{tabular}
\end{table}

\begin{figure}[t!]
    \centering
    \includegraphics[width=0.8\columnwidth]{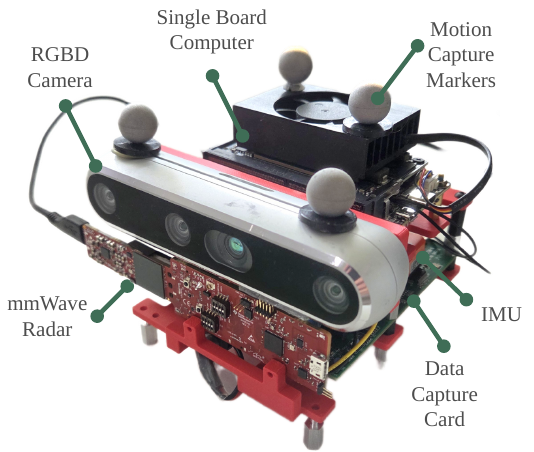}
    \caption{Sensor acquisition rig comprising a TI AWR6843AOP-EVM mmWave FMCW radar,
DCA1000 data-capture card, NVIDIA Jetson Orin NX single-board computer, and
Xsens MTi-320-3A IMU. Retro-reflective markers provide indoor motion-capture
ground truth. (RGB-D camera not used in this evaluation.)}
    \label{fig:rig}
    \Description{Sensor acquisition rig. The rig contains TI AWR6843AOP-EVM mmWave FMCW
    radar, DCA1000 data-capture card, NVIDIA
    Jetson Orin NX single-board computer, Xsens MTi-320-3A IMU.
    Retro-reflective markers enable MoCap ground truth indoors. }
\end{figure}

\begin{figure}[t!]
  \centering
  \resizebox{\columnwidth}{!}{%
  \begin{tikzpicture}[
      font=\footnotesize,
      >={Stealth[length=1.6mm]},
      line width=0.5pt,
      io/.style={draw,rounded corners,align=center,inner sep=2pt,
                 minimum height=7mm,text width=13mm},
      m/.style={draw,rounded corners,align=center,inner sep=2pt,
                minimum height=7mm,text width=24mm},
      sol/.style={draw,rounded corners,align=center,inner sep=2pt,
                  minimum height=7mm,text width=23mm},
      ours/.style={fill=OIGreen!22},
      pose/.style={draw,rounded corners,align=center,inner sep=2.5pt,
                   minimum height=9mm,text width=15mm},
    ]
    \node[io] (tichip) at (0,0)    {TI on-chip\\cloud};
    \node[io] (rawadc) at (0,-3.0) {Raw ADC\\cube};
    \node[m] (onb) at (2.45,0)     {\texttt{ptcloud\_onboard}};
    \node[m] (fft) at (2.45,-1.15) {\texttt{ptcloud\_dsp\_fft}\\ host CFAR, FFT};
    \node[m] (cap) at (2.45,-2.30) {\texttt{ptcloud\_dsp\_capon}\\ host CFAR, Capon};
    \node[m, ours] (dsw) at (2.45,-4.05) {\textbf{Dense Soft Weighting}};
    \node[m] (learn) at (2.45,-5.95)
          {\textbf{deep network}\\ native trained model};
    \node[sol, text width=19mm] (rsac) at (5.1,-1.15) {RANSAC +\\Least Squares};
    \node[sol, ours] (wls) at (5.1,-4.05) {\textbf{Robust Weighted}\\\textbf{Least Squares}};
    \node[draw,rounded corners,align=center,inner sep=3pt,
          minimum height=32mm,text width=13mm] (eskf) at (7.55,-2.6)
          {\textbf{Shared}\\\textbf{ESKF}\\[2pt] identical\\parameters};
    \node[pose] (pose) at (9.25,-2.6) {Pose /\\Trajectory};
    \node[font=\footnotesize\itshape, align=right, anchor=north east]
          at ([xshift=-1mm,yshift=-0.5mm]pose.south) {eval.\\output};
    \node[font=\large\bfseries] (ttl) at (4.7,1.5) {Evaluation setup};
    \draw[line width=0.6pt] (ttl.south west) -- (ttl.south east);
    \draw[->] (tichip) -- (onb);                         %
    \draw (rawadc.east) -- (0.95,-3.0);                   %
    \draw[->] (0.95,-3.0) |- (fft.west);
    \draw[->] (0.95,-3.0) |- (cap.west);
    \draw[->] (0.95,-3.0) |- (dsw.west);
    \draw[->] (0.95,-3.0) |- (learn.west);
    \draw (onb.east) -- (3.95,0);
    \draw (fft.east) -- (3.95,-1.15);
    \draw (cap.east) -- (3.95,-2.30);
    \draw (3.95,0) -- (3.95,-2.30);                       %
    \draw[->] (3.95,-1.15) -- (rsac.west);
    \draw[->] (dsw) -- (wls);
    \draw[dashed,black!55] (6.5,-0.55) -- (6.5,-4.75);
    \node[font=\footnotesize\itshape, anchor=south] at (6.5,-0.45)
          {$\hat{\bm{v}}_{\text{sensor}},\,\bm{\Sigma}_{\text{sensor}}$};
    \draw[->] (rsac.east) -- (eskf.west |- rsac);
    \draw[->] (wls.east)  -- (eskf.west |- wls);
    \draw[->] (eskf.east) -- (pose.west);
    \draw[->] (learn.south) -- (2.45,-6.85) -- (9.25,-6.85) -- (pose.south);
    \node[font=\footnotesize\itshape, above] at (6.0,-6.85)
          {learning-based: direct pose};
    \begin{scope}[on background layer]
      \node[draw, dashed, rounded corners, fit=(onb)(cap), inner sep=1.6mm,
            label={[font=\footnotesize\itshape]above:Point-cloud baselines (indirect)}] {};
      \node[draw, dashed, rounded corners, fit=(dsw)(wls), inner sep=2mm,
            label={[font=\footnotesize\itshape\bfseries]above:this work}] {};
      \node[draw, dashed, rounded corners, fit=(learn), inner sep=1.6mm,
            label={[font=\footnotesize\itshape]above:Learning-Based RIO baselines}] {};
    \end{scope}
  \end{tikzpicture}%
  }
\caption{Cross-method comparison harness (left to right); only
    \emph{this work} (DSW and its robust WLS) is shaded. The model-based
    paths hand the back-end the same
    $(\hat{\bm{v}}_{\text{sensor}}, \bm{\Sigma}_{\text{sensor}})$ across the
    dashed interface, and the shared ESKF uses identical parameters, so
    trajectory differences attribute to the front-end alone. The learning-based
    baseline regresses pose directly, bypassing this interface and the
    ESKF.}
  \label{fig:harness}
  \Description{Left-to-right block diagram of the evaluation harness. Two
    input boxes on the left, a raw ADC cube and a TI on-chip point cloud,
    feed a stacked column of front-ends. Three point-cloud baselines (the
    on-chip cloud feeds ptcloud_onboard; the raw ADC cube feeds the two
    host-CFAR baselines) share a single RANSAC + LS estimator; DSW, the
    contribution of this work, operates directly on the dense cube and
    uses robust WLS, and DSW and its estimator are the only shaded boxes.
    Both model-based paths hand the same sensor-frame velocity and covariance
    across a dashed interface line into one tall shared error-state Kalman
    filter, which produces a pose/trajectory output on the right. A learning-based
    radar-inertial network (Radarize or milliEgo) fed from the raw ADC cube
    regresses pose directly along the bottom, bypassing the velocity
    interface and the filter, and is compared on the same pose output.}
\end{figure}
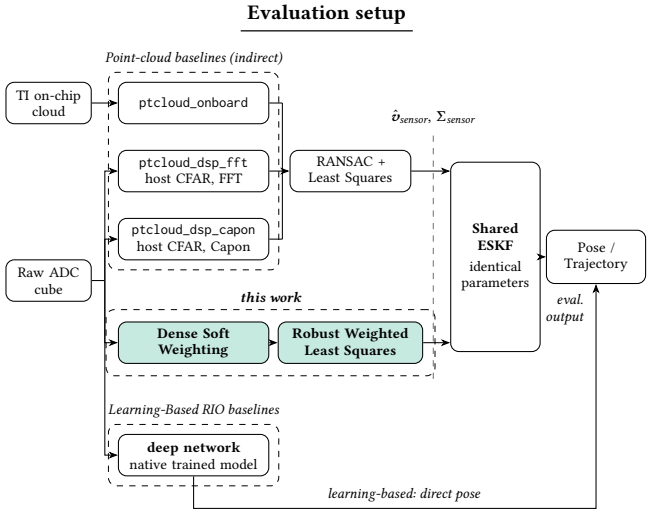

\begin{figure*}[t!]
    \centering
    \includegraphics[width=\textwidth]{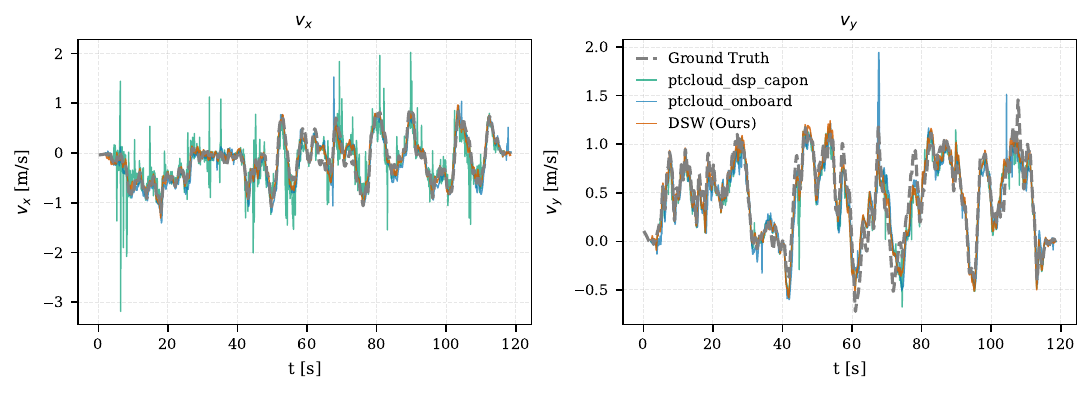}
    \caption{Per-frame body-frame $v_x$ and $v_y$ over time on one
    ColoRadar sequence \texttt{arpg\_lab\_run1}. The proposed system matches the
    ground-truth velocity most closely, attaining the lowest per-frame
    error among the compared methods.}
    \label{fig:velocity-arpg}
    \Description{Per-frame body-frame x and y ego-velocity over time on one
    ColoRadar sequence; The proposed system matches the ground-truth velocity most closely,
    attaining the lowest per-frame error among the compared methods.}
\end{figure*}

Our evaluation examines the dense ego-velocity front-end from four angles.
We first ask whether dense soft weighting improves per-frame velocity and
trajectory accuracy relative to sparse point-cloud front-ends
(\cref{sec:results-velocity,sec:results-coloradar}). We then isolate which
weighting, robustness, and covariance components drive the fused odometry
accuracy (\cref{sec:ablation}). Next, we test whether the same fixed
front-end transfers across radar chirp configurations, devices, and a
comparison with learning-based methods without retuning
(\cref{sec:results-selfcollected,sec:results-radarize}). Finally, we measure
whether dense ego-velocity estimation remains practical on desktop and
embedded compute targets (\cref{sec:runtime}).

\subsection{Experimental Setup}
\label{sec:datasets}

\paragraph*{Datasets.}
We evaluate the dense ego-velocity front-end on three mmWave FMCW radar
datasets. All three use TI single-chip radars and provide DCA1000 raw
ADC captures, which dense soft weighting requires because it operates on
the full range-Doppler cube rather than on detected point clouds.
\Cref{tab:datasets} summarises the sensing platforms, ground-truth
sources, and chirp configurations. Unless otherwise stated, all
front-end hyperparameters are fixed across datasets and sequences
(\cref{sec:baselines}).

\textbf{ColoRadar Dataset}~\cite{Kramer2022ColoRadar} contains $52$ handheld
sequences totalling more than \SI{145}{\minute} of radar, lidar, and IMU
data from a single-chip AWR1843BOOST with DCA1000 capture, recorded in
indoor and outdoor environments.

Each sequence provides both raw ADC captures and on-chip DSP point
clouds from the same radar frames, so the dense front-end and the
sparse point-cloud baselines run on identical recordings. The dataset
provides two ground-truth sources: a lidar-SLAM trajectory for every
sequence and motion-capture poses for the sequences recorded in the
ASPEN laboratory. Because the evaluated sequences span indoor and
outdoor environments, we use the lidar-SLAM reference for all of them.
We use three sequences, covering
\SI{336.8}{\second} and \SI{300.34}{\meter} in the cached ground-truth
trajectories; these forward-driving, predominantly planar sequences are
the longest in our evaluation.

\textbf{Radarize dataset.} The second dataset is the evaluation data
released with the Radarize radar-SLAM system~\cite{Radarize2024},
comprising $146$ indoor handheld, cart, and robot sequences acquired
with an AWR1843BOOST radar and an Intel RealSense T265. We select it
because it provides raw ADC captures and is the native training and
evaluation setting of the learning-based radar odometry baselines,
making it the closest available basis for comparing the dense analytic
front-end against learned methods. The dataset provides no dedicated
ground-truth system or standalone IMU: following the original protocol,
we use the T265 visual-inertial trajectory as pseudo-ground truth and
the T265's internal IMU as the inertial input, and all compared methods
are evaluated against the same locally associated pseudo-ground-truth
poses. The held-out split contains $89$ sequences totalling
\SI{92.9}{\minute} and \SI{3.28}{\kilo\meter} in the pseudo-ground-truth
trajectories.

\textbf{Self-collected dataset.} The third dataset was recorded using our own sensor rig (\cref{sec:self-hardware}, \cref{fig:rig}), which also defines the embedded deployment class used in the runtime study. Unlike the two public datasets, this dataset uses a different radar platform and configuration: the antenna-on-package AWR6843AOP. Although both the AWR1843BOOST and AWR6843AOP use twelve-element virtual arrays, the two antenna layouts produce different effective apertures: the AWR1843BOOST configurations used in ColoRadar and Radarize are optimized primarily for azimuth, whereas the AWR6843AOP provides a more balanced azimuth--elevation field of view. Combined with strongly three-dimensional handheld motion, this configuration explicitly exercises the elevation axis, which remains largely planar in the public benchmarks.

The rig records radar and IMU data together, and all sequences were collected inside a motion-capture volume to obtain reference poses. The dataset comprises $50$ sequences, totalling approximately ${\sim}30{,}000$ radar frames at a frame rate of \SI{10}{\hertz}, together with motion-capture ground truth. The sequences span a range of three-dimensional handheld motions.
\subsection{Self-Collected Hardware Platform}
\label{sec:self-hardware}

For self-collected rig (\cref{fig:rig}), we use an NVIDIA Jetson Orin NX single-board computer, a TI AWR6843AOP-EVM radar, a DCA1000 capture
card, and an Xsens MTi-320-3A AHRS. The AOP-EVM virtual array provides a
$60^\circ/60^\circ$ azimuth/elevation field of view
(\cref{tab:datasets}), substantially wider in elevation than the
AWR1843BOOST apertures used by the public
datasets. This configuration is the main reason the self-collected
sequences exercise full 3D ego-velocity rather than primarily planar
motion.

Ground truth is provided by a 10-camera Vicon Vero v2.2 system tracking
retro-reflective markers on the rig; the Vicon trajectory is aligned to
the estimator clock using an offline per-trajectory time-offset
estimate~\cite{Kim2025EKFRIOTC}. The same platform motivates the
embedded-runtime measurements in \cref{sec:runtime}: the front-end must
run within the \SI{10}{\hertz} frame budget on a deployable compute
target, not only on a development laptop.

Runtime measurements use two compute platforms. The development laptop
has an Intel Core i9-14900HX CPU, \SI{64}{\giga\byte} of RAM, and an
NVIDIA RTX~4090 Mobile GPU. The embedded target uses an NVIDIA Jetson
Orin~NX Module 16GB on a Seeed reComputer Mini Carrier Board; the module
has an 8-core Arm Cortex-A78AE CPU, \SI{16}{\giga\byte} of LPDDR5 memory,
and a 1024-core NVIDIA Ampere GPU with 32 Tensor Cores.

\subsection{Baselines}
\label{sec:baselines}

We compare Dense Soft Weighting against two families of baselines:
analytic point-cloud ego-velocity pipelines and learning-based radar-odometry
networks. \Cref{fig:harness} illustrates how both families connect to the
shared inertial back-end. The setup is designed so that, for the analytic
methods, the radar front-end is the only component that varies. All
front-ends consume the same recordings: the raw ADC cube feeds the
host-CFAR baselines, the dense front-end, and the learning-based networks, while
\ptcloudonboard{} consumes the TI on-chip point cloud distributed with the
same frames. Every model-based front-end, sparse or dense, outputs a
sensor-frame ego-velocity estimate and covariance; the dashed line in
\cref{fig:harness} marks this common interface.

Downstream of this interface, every model-based front-end feeds the
same loose-coupled ESKF~(\cref{sec:eskf}) with identical parameters and
the same process-noise matrix $\Qmat$, the same online radar--IMU
temporal-offset state $t_d$, and the same $\chi^2_3$ outlier gate. We
apply no per-method or per-sequence filter tuning; the exact settings are
listed under Reproducibility. Trajectory-level differences therefore
attribute to the front-end, not to filter design. The learning-based baselines
(\cref{sec:results-radarize}) regress pose directly from the radar stream
and bypass both the velocity interface and the filter; they are compared
at the pose level on the same sequences and metrics.

\paragraph*{Point-cloud baselines}
We compare the dense front-end with sparse, point-cloud-based
ego-velocity pipelines that follow the RANSAC family~\cite{Doer2020rio, Doer2021xRIO,
Kim2025EKFRIOTC}: 3-point RANSAC rejects gross Doppler outliers, after
which a least-squares fit to the consensus set, refined by orthogonal
distance regression when enough inliers survive, recovers the
ego-velocity. The three point-cloud variants differ only in
how the cloud is detected (\cref{tab:ate}): \ptcloudonboard{} reuses the
TI on-chip cloud distributed with the frames, whereas \ptclouddspcapon{}
and \ptclouddspfft{} are detected host-side from the same raw ADC frames
with Capon digital beamforming~\cite{Jiang2025DBE} and FFT/Bartlett
angle-of-arrival estimation, respectively. On ColoRadar we compare
against \ptcloudonboard{} and \ptclouddspcapon{}; on the self-collected
sequences the on-chip cloud is unavailable, so we substitute
\ptclouddspfft{} for \ptcloudonboard{}. All point-cloud baselines and the
dense front-end feed the same ESKF described above.

\paragraph*{Learning-Based baselines}
On Radarize, we compare at the pose level with Radarize~\cite{Radarize2024},
a Doppler-flow network, and milliEgo~\cite{milliEgo2020}, a radar-inertial
ego-motion network. Both use the Radarize radar stream and dataset
protocol and are evaluated on the same held-out split as the dense
front-end.

\paragraph*{Reproducibility}
We fix every front-end hyperparameter across all sequences and report
them here so the comparison can be reproduced. The point-cloud detector
applies cascaded cell-averaging smallest-of CFAR, an adaptive threshold with
binary output, to the range-Doppler power map (range axis, then Doppler axis)
at a false-alarm rate of
$10^{-2}$ on each axis, with $8$ guard and $8$ training cells along
range, $0$ guard and $4$ training cells along Doppler, and peak grouping
disabled; surviving detections are projected over an azimuth/elevation
field of view of $60^\circ/15^\circ$ on the AWR1843BOOST configurations
used by ColoRadar and Radarize, and $60^\circ/60^\circ$ on the
AWR6843AOP-EVM self-collected configuration. The sparse estimator draws
$100$ minimal three-point hypotheses,
counts a cell as an inlier when its Doppler residual falls below
\SI{0.15}{\meter\per\second}, requires a three-point consensus, performs
unweighted least squares on the inlier set, and refines with orthogonal
distance regression whenever at least six inliers remain; its covariance
is the residual sandwich floored by the geometric term
$\sigma_d^2(\mat{U}^\transp\mat{U})\inv$. The dense front-end fixes the
gate parameters $\tau=200$ and $\kappa=0.5$ (\cref{eq:weight}), takes a
single Cauchy pass with a fixed cutoff $c=2$ on both apertures
(\cref{sec:wls}), and adds only a per-platform scalar floor to
the covariance (\cref{sec:fisher}). Both front-ends feed the identical ESKF,
whose IMU process noise follows EKF-RIO~\cite{Kim2025EKFRIOTC}
(accelerometer and gyroscope noise densities of $0.03~\mathrm{m/s^2}$ and
$0.2~^{\circ}/\mathrm{s}$ per $\sqrt{\mathrm{Hz}}$, bias random walks of
$10^{-5}$ in SI units) under a $\chi^2_3$ innovation gate at the $0.95$
confidence level.

\subsection{Metrics and Protocol}
Absolute pose error (APE) and relative pose error (RPE) are computed with
evo~\cite{Grupp2017evo}. For ColoRadar and the self-collected dataset,
the cached evaluator uses evo-compatible SE(3) origin alignment
(\alignoriginflag{}); map overlays use Umeyama SE(3) alignment only
for visualisation. ColoRadar uses a \SI{10}{\meter}. 
Radarize is evaluated with Umeyama alignment after \SI{0.01}{\second}
trajectory association and native per-frame RPE ($\Delta=1$ frame),
which is retained for comparability with its published evaluation
protocol and learning-based baselines. Per-axis velocity RMSE is computed from
time-associated body-frame velocity samples after the same offline
ground-truth time alignment used for the pose trajectory.

\section{Results}
\label{sec:results}

\begin{figure*}[tp]
    \centering
    \subfloat[\texttt{arpg\_lab\_run0}]{%
        \includegraphics[width=0.216\textwidth]{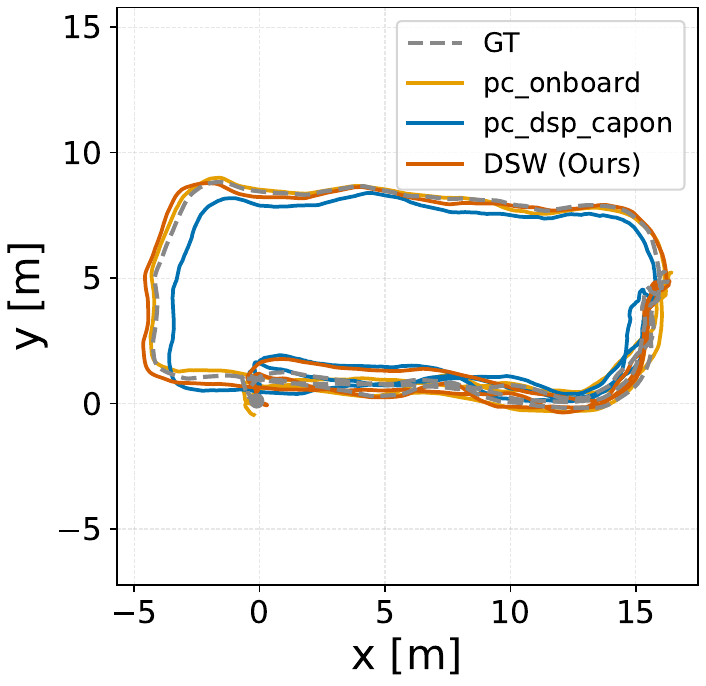}}\hfill
    \subfloat[\texttt{arpg\_lab\_run1}]{%
        \includegraphics[width=0.216\textwidth]{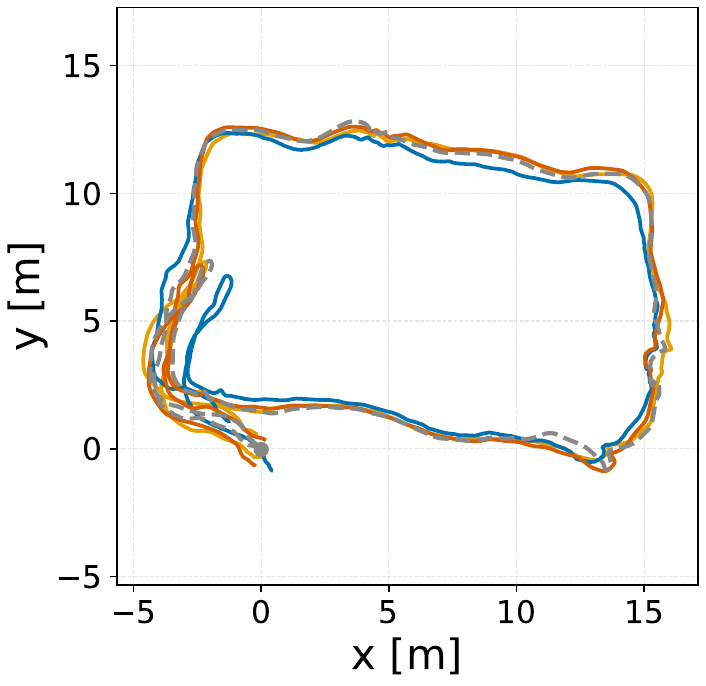}}\hfill
    \subfloat[\texttt{ec\_hallways\_run0}]{%
        \includegraphics[width=0.216\textwidth]{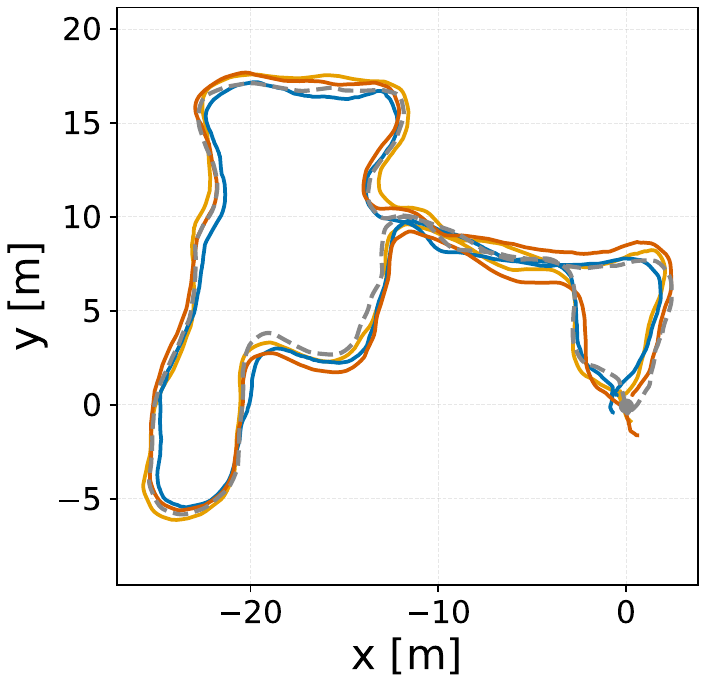}}\hfill
    \subfloat[{ColoRadar $\overline{\text{APE}}_t$ \& $\overline{\text{RPE}}_t$ [m]}]{%
        \raisebox{0.1\height}{\includegraphics[width=0.216\textwidth]{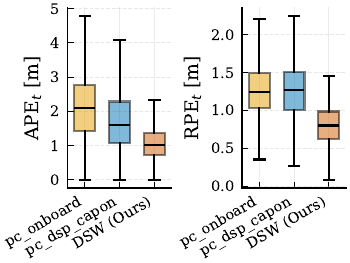}}}

    \vspace{0.3em}

    \subfloat[\texttt{cart\_csl\_upper\_loop\_0}]{%
        \includegraphics[width=0.216\textwidth]{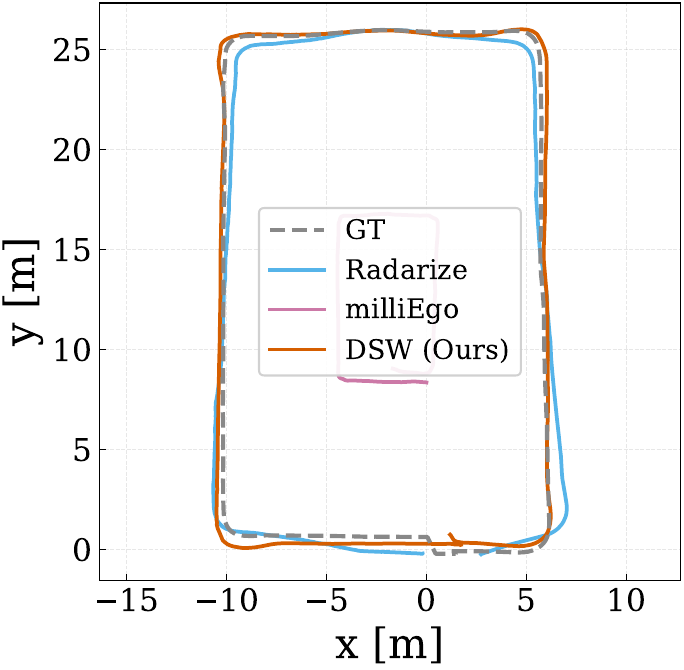}}\hfill
    \subfloat[\texttt{cart\_csl\_upper\_2}]{%
        \includegraphics[width=0.216\textwidth]{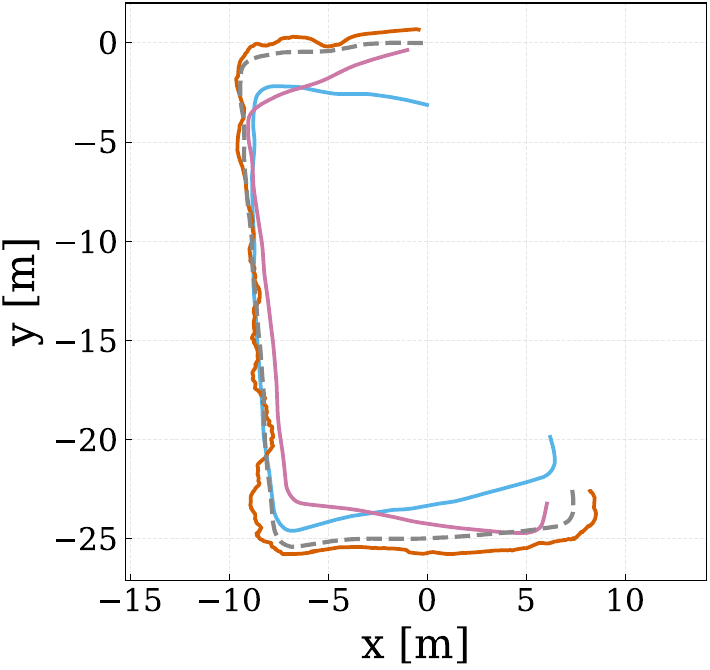}}\hfill
    \subfloat[\texttt{walk\_sc\_basement\_0}]{%
        \includegraphics[width=0.216\textwidth]{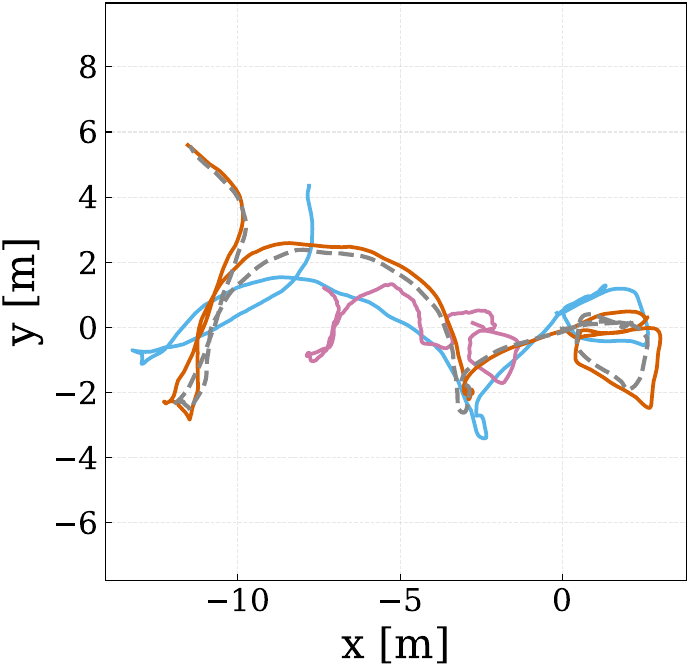}}\hfill
    \subfloat[{Radarize set $\overline{\text{APE}}_t$ \& $\overline{\text{RPE}}_t$ [m]}]{%
        \raisebox{0.1\height}{\includegraphics[width=0.216\textwidth]{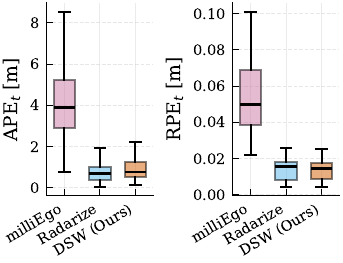}}}

    \vspace{0.3em}

    \subfloat[\texttt{sequence\_01}]{%
        \includegraphics[width=0.216\textwidth]{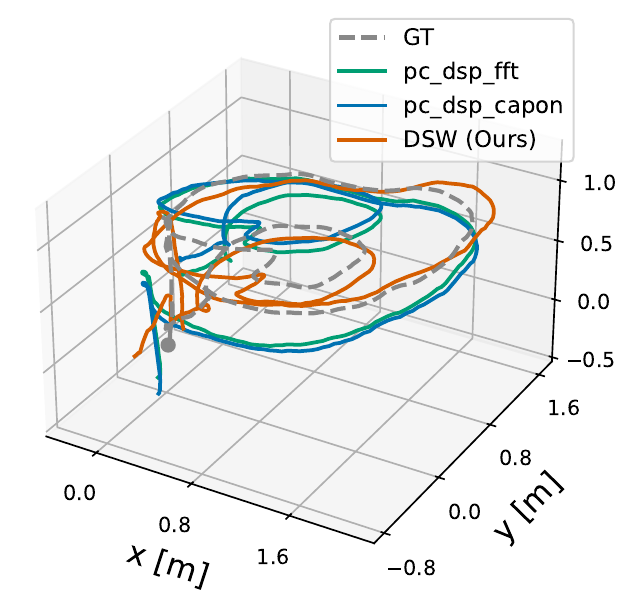}}\hfill
    \subfloat[\texttt{sequence\_02}]{%
        \includegraphics[width=0.216\textwidth]{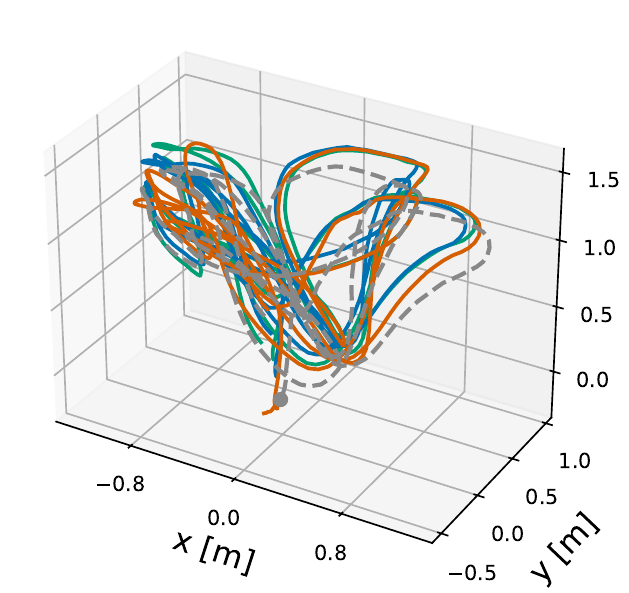}}\hfill
    \subfloat[\texttt{sequence\_03}]{%
        \includegraphics[width=0.216\textwidth]{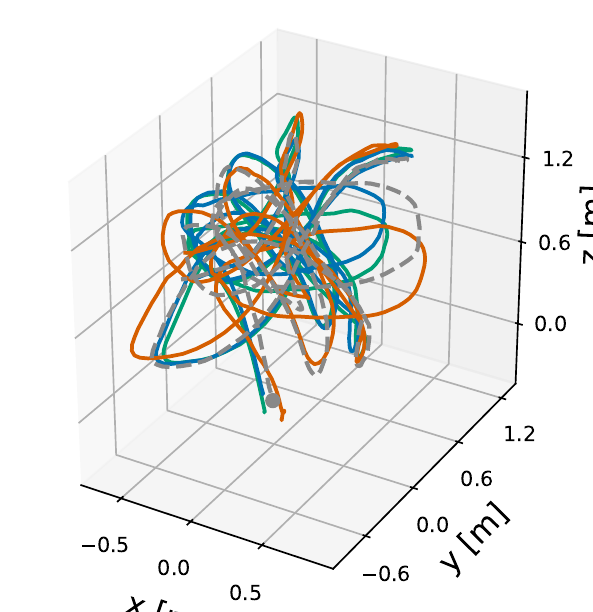}}\hfill
    \subfloat[{Self-collected $\overline{\text{APE}}_t$ \& $\overline{\text{RPE}}_t$ [m]}]{%
        \raisebox{0.1\height}{\includegraphics[width=0.216\textwidth]{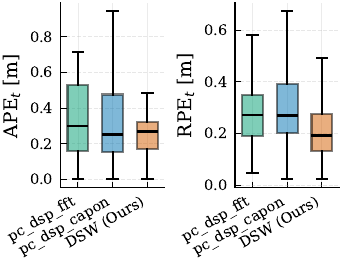}}}
    \caption{Estimated trajectories across the three evaluation datasets:
    ColoRadar (top, XY view), Radarize (middle), and
    the self-collected rig (bottom, XYZ view). The first three panels in
    each row show representative sequences; the fourth panel summarises
    APE\textsubscript{t} and RPE\textsubscript{t}.}
    \label{fig:trajectories}
    \Description{Estimated trajectories across three datasets, one row each:
    ColoRadar, the Radarize dataset, and the self-collected rig, with per-dataset
    error box plots; The proposed system tracks ground truth closely, matching the learning-based
    Radarize network and improving on milliEgo and the point-cloud baselines.}
\end{figure*}

\begin{table*}[t]
    \centering
    \caption{Per-sequence ego-velocity and trajectory error on ColoRadar,
    Radarize, and the self-collected dataset.}
    \label{tab:ate}
    \renewcommand{\arraystretch}{0.95}
    \small
    \begin{tabular*}{0.90\textwidth}{@{\extracolsep{\fill}}l l l l rrrrr@{}}
    \toprule
     & & \multicolumn{2}{c}{\textbf{Baseline}} & \multicolumn{1}{c}{$\Delta v$} & \multicolumn{2}{c}{\textbf{APE}} & \multicolumn{2}{c}{\textbf{RPE}} \\
    \cmidrule(lr){3-4}\cmidrule(lr){5-5}\cmidrule(lr){6-7}\cmidrule(l){8-9}
    \textbf{Dataset} & \textbf{Sequence} & \textbf{Method}\textsuperscript{$\dagger$} & \textbf{Family} & [m/s] & t [m] & r [$^\circ$] & t [m] & r [$^\circ$] \\
    \midrule
    \multirow{9}{*}{ColoRadar}
      & \multirow{3}{*}{arpg\_lab\_run0} & ptcloud\_onboard    & Point-cloud & 0.54 & 2.16 & \textbf{1.56} & 1.40 & \textbf{1.40} \\
      & & ptcloud\_dsp\_capon & Point-cloud & 0.35 & 1.48 & 1.37 & 1.51 & 1.61 \\
      & & \textbf{DSW (ours)} & Direct & \textbf{0.30} & \textbf{0.86} & 1.36 & \textbf{0.85} & 1.56 \\
    \cmidrule(l){2-9}
      & \multirow{3}{*}{arpg\_lab\_run1} & ptcloud\_onboard    & Point-cloud & 0.34 & 2.02 & 2.36 & 1.36 & 3.11 \\
      & & ptcloud\_dsp\_capon & Point-cloud & 0.51 & 1.66 & 2.45 & 1.25 & 2.03 \\
      & & \textbf{DSW (ours)} & Direct & \textbf{0.32} & \textbf{1.04} & \textbf{2.08} & \textbf{1.04} & \textbf{1.91} \\
    \cmidrule(l){2-9}
      & \multirow{3}{*}{ec\_hallways\_run0} & ptcloud\_onboard    & Point-cloud & 0.45 & 2.88 & 2.29 & 1.38 & 1.57 \\
      & & ptcloud\_dsp\_capon & Point-cloud & 0.57 & 3.21 & \textbf{1.99} & 1.52 & 1.53 \\
      & & \textbf{DSW (ours)} & Direct & \textbf{0.34} & \textbf{1.60} & 2.22 & \textbf{0.88} & \textbf{1.44} \\
    \midrule
    \multirow{9}{*}{Radarize}
      & \multirow{3}{*}{cart\_csl\_upper\_loop\_0} & milliEgo & Learning-Based & -- & 8.17 & 12.3 & 0.083 & 0.29 \\
      & & Radarize & Learning-Based & -- & 0.66 & 7.1 & 0.026 & 3.62 \\
      & & \textbf{DSW (ours)} & Direct          & 0.24 & \textbf{0.29} & \textbf{4.3} & \textbf{0.021} & \textbf{0.28} \\
    \cmidrule(l){2-9}
      & \multirow{3}{*}{cart\_csl\_upper\_2} & milliEgo & Learning-Based & -- & 1.60 & 172.2 & 0.163 & 0.37 \\
      & & Radarize & Learning-Based & -- & 1.38 & \textbf{6.4} & \textbf{0.019} & 1.68 \\
      & & \textbf{DSW (ours)} & Direct          & 0.08 & \textbf{0.55} & 6.5 & 0.022 & \textbf{0.29} \\
    \cmidrule(l){2-9}
      & \multirow{3}{*}{walk\_sc\_basement\_0} & milliEgo & Learning-Based & -- & 3.10 & 16.7 & 0.049 & 0.95 \\
      & & Radarize & Learning-Based & -- & 1.21 & 28.4 & 0.016 & 3.25 \\
      & & \textbf{DSW (ours)} & Direct          & 0.09 & \textbf{0.40} & \textbf{8.0} & \textbf{0.015} & \textbf{0.73} \\
    \midrule
    \multirow{9}{*}{Self-collected}
      & \multirow{3}{*}{sequence\_01} & ptcloud\_dsp\_fft   & Point-cloud & 0.16 & 0.37 & 3.24 & \textbf{0.10} & \textbf{2.35} \\
      & & ptcloud\_dsp\_capon & Point-cloud & 0.12 & 0.48 & 3.22 & 0.11 & 2.44 \\
      & & \textbf{DSW (ours)} & Direct & \textbf{0.08} & \textbf{0.28} & \textbf{3.20} & 0.16 & 2.61 \\
    \cmidrule(l){2-9}
      & \multirow{3}{*}{sequence\_02} & ptcloud\_dsp\_fft   & Point-cloud & 0.12 & 0.34 & 3.02 & \textbf{0.08} & 2.41 \\
      & & ptcloud\_dsp\_capon & Point-cloud & 0.12 & 0.29 & 3.04 & 0.08 & 2.51 \\
      & & \textbf{DSW (ours)} & Direct & \textbf{0.10} & \textbf{0.27} & \textbf{2.92} & 0.10 & \textbf{2.33} \\
    \cmidrule(l){2-9}
      & \multirow{3}{*}{sequence\_03} & ptcloud\_dsp\_fft   & Point-cloud & 0.14 & 0.45 & 3.51 & 0.11 & 3.19 \\
      & & ptcloud\_dsp\_capon & Point-cloud & 0.14 & 0.39 & 3.40 & 0.11 & 3.35 \\
      & & \textbf{DSW (ours)} & Direct & \textbf{0.12} & \textbf{0.25} & \textbf{2.90} & \textbf{0.09} & \textbf{2.88} \\
    \bottomrule
    \end{tabular*}

    \smallskip
    {\footnotesize\renewcommand{\arraystretch}{1.0}%
    \begin{minipage}{0.90\textwidth}\raggedright
    \textsuperscript{$\dagger$}``--'' = not available; best per sequence and
    metric in \textbf{bold}.\\[2pt]
    \textbf{Metrics.} $\Delta v$ is per-frame body-velocity RMSE;
    APE\textsubscript{t/r} and RPE\textsubscript{t/r} are absolute and
    relative pose error (translation\,/\,rotation).\\[2pt]
    \textbf{Methods.}
    \emph{Point-cloud} (host CFAR detection $\rightarrow$ RANSAC\,+\,LS\,+\,ODR):
    \ptcloudonboard{} TI on-chip DSP cloud;
    \ptclouddspfft{} FFT/Bartlett AoA cloud;
    \ptclouddspcapon{} Capon digital-beamforming cloud (DBE~\cite{Jiang2025DBE}).
    \emph{Direct}: DSW (ours) dense range-Doppler soft weighting + robust WLS.
    \emph{Learning-based} (end-to-end pose regression): Radarize~\cite{Radarize2024}
    Doppler-flow network; milliEgo~\cite{milliEgo2020} radar-inertial network.
    All analytic methods share one ESKF.
    \end{minipage}\par}
\end{table*}
\begin{table}[t]
  \centering
  \begin{threeparttable}
  \caption{Per-frame latency (median, \si{\milli\second}) by dataset and
  platform.}
  \label{tab:runtime}
  \renewcommand{\arraystretch}{0.90}
  \setlength{\tabcolsep}{3pt}
  \footnotesize
  \begin{tabular}{@{}l r l l rrr@{}}
  \toprule
  & & & & \multicolumn{2}{c}{\textbf{CPU cores}} & \\
  \cmidrule(lr){5-6}
  \textbf{Dataset}
    & \makecell[c]{\textbf{RD}\\[-1pt]\textbf{cells}}
    & \textbf{Device} & \textbf{Method}
    & \textbf{1$\times$} & \textbf{4$\times$} & \textbf{GPU} \\
  \midrule
  \multirow{6}{*}{ColoRadar} & \multirow{6}{*}{\num{16384}}
    & \multirow{3}{*}{Laptop}
                        & DSW (ours)         & 17.8 & 8.8  & 2.0 \\
  & & & \ptclouddspfft{}    & 10.1 & 6.6  & 3.2 \\
  & & & \ptclouddspcapon{}  & 12.3 & 9.7  & 4.1 \\
  \cmidrule(l){3-7}
  & & \multirow{3}{*}{Orin~NX}
                        & DSW (ours)         & 71   & 35   & 12  \\
  & & & \ptclouddspfft{}    & 40   & 26   & 19  \\
  & & & \ptclouddspcapon{}  & 49   & 39   & 25  \\
  \midrule
  \multirow{6}{*}{\makecell[l]{Self-\\collected}} & \multirow{6}{*}{\num{15300}}
    & \multirow{3}{*}{Laptop}
                        & DSW (ours)         & 17.7 & 8.9  & 2.1 \\
  & & & \ptclouddspfft{}    & 10.4 & 6.9  & 4.0 \\
  & & & \ptclouddspcapon{}  & 13.5 & 10.4 & 4.9 \\
  \cmidrule(l){3-7}
  & & \multirow{3}{*}{Orin~NX}
                        & DSW (ours)         & 71   & 36   & 13  \\
  & & & \ptclouddspfft{}    & 42   & 28   & 24  \\
  & & & \ptclouddspcapon{}  & 54   & 42   & 29  \\
  \midrule
  \multirow{10}{*}{Radarize} & \multirow{10}{*}{\num{3072}}
    & \multirow{5}{*}{Laptop}
                        & DSW (ours)         & 3.8  & 1.9  & 1.8 \\
  & & & \ptclouddspfft{}    & 3.9  & 2.6  & 3.1 \\
  & & & \ptclouddspcapon{}  & 4.9  & 3.8  & 3.8 \\
  & & & Learning-Based (Radarize) & 54.9 & 32.8 & 3.4 \\
  & & & Learning-Based (milliEgo) & 20.2 & 11.7 & 1.2 \\
  \cmidrule(l){3-7}
  & & \multirow{5}{*}{Orin~NX}
                        & DSW (ours)         & 15   & 8    & 11  \\
  & & & \ptclouddspfft{}    & 16   & 11   & 19  \\
  & & & \ptclouddspcapon{}  & 20   & 15   & 23  \\
  & & & Learning-Based (Radarize) & 220  & 134  & 21  \\
  & & & Learning-Based (milliEgo) & 81   & 48   & 7   \\
  \bottomrule
  \end{tabular}
  \begin{tablenotes}[flushleft]
  \footnotesize
  \item[] Laptop and Orin~NX specifications are given in
  \cref{sec:self-hardware}.
  \item[] RD cells is the dataset's range-Doppler FFT cell count per frame.
  CPU columns are core counts; GPU uses CUDA.
  \item[] Learning-Based rows are the per-frame network inference of the
  learned baselines (PyTorch FP32)
  \end{tablenotes}
  \end{threeparttable}
\end{table}

\begin{table}[t]
  \centering
  \begin{threeparttable}
  \caption{Ablation of the front-end design choices, mean over the six
  primary sequences.}
  \label{tab:ablation}
  \renewcommand{\arraystretch}{0.95}
  \setlength{\tabcolsep}{4pt}
  \small
  \begin{tabularx}{0.95\columnwidth}{@{}>{\raggedright\arraybackslash}Xrr@{}}
  \toprule
   & \multicolumn{2}{c}{\textbf{Mean error}} \\
  \cmidrule(l){2-3}
  \textbf{Variant} & APE\textsubscript{t} [m] & $\Delta v$ [m/s] \\
  \midrule
  DSW (default)                       & 0.76 & 0.23 \\
  \quad power score only                  & 2.47 & 0.32 \\
  \quad concentration gate only           & 2.97 & 0.36 \\
  \quad hard gate (vs.\ soft)             & diverges & -- \\
  \quad scalar $\bm\Sigma_v$ (vs.\ 3-component) & 0.79 & 0.23 \\
  \quad $\tau{=}100$ / $400$ (def.\ 200)  & 0.77 / 0.77 & -- \\
  \quad $\kappa{=}0.25$ / $1.0$ (def.\ 0.5) & 1.11 / 1.02 & -- \\
  \bottomrule
  \end{tabularx}
  \begin{tablenotes}[flushleft]
  \footnotesize
  \item[] Six primary sequences: three ColoRadar + three self-collected.
  Each row toggles one element from the default and re-runs the shared ESKF.
  APE\textsubscript{t} is translational absolute pose error, $\Delta v$ the
  per-frame body-velocity RMSE. The hard-gate variant diverges (mean
  APE\textsubscript{t} $>$\SI{200}{\meter}).
  \end{tablenotes}
  \end{threeparttable}
\end{table}

\subsection{Per-Frame Ego-Velocity Accuracy}
\label{sec:results-velocity}

Trajectory accuracy is reported after inertial integration. However, the
front-end contribution appears one step earlier in the per-frame
body-velocity estimate. \Cref{fig:velocity-arpg} compares the two
horizontal velocity components on a representative ColoRadar sequence.
The dense estimate matches the reference most closely and attains the
lowest per-frame error; the Capon point-cloud baseline follows the same
trend with higher error, and the on-chip point-cloud baseline has the
highest. The
$\Delta v$ column of \cref{tab:ate} reports the same quantity over all
sequences, and dense soft weighting attains the lowest per-frame
ego-velocity RMSE on every ColoRadar row.

\Cref{fig:weighting}(c--e) links this velocity improvement to the
front-end representation. CASO-CFAR adapts its threshold to the local
range-Doppler neighbourhood. However, its output remains a sparse,
boresight-biased binary detection set. The dense front-end assigns a
continuous weight to
every range-Doppler cell and retains off-boresight evidence that the
detector removes. Under forward motion the boresight detections are often
sufficient to constrain $v_x$; under lateral motion the off-boresight
cells are the measurements that constrain $v_y$. The improvement is
therefore present before the ESKF receives the measurement, rather than
being introduced by back-end tuning.

\subsection{Front-End Component Ablation}
\label{sec:ablation}

We ablate the front-end on the six primary
sequences by toggling one element at a time and re-running the shared
ESKF (\cref{tab:ablation}). The two factors in~\cref{eq:weight} are both
necessary: using only the power score or only the concentration gate
raises mean APE\textsubscript{t} to $2.47$~m and $2.97$~m, against
$0.76$~m for the full weight. Replacing the soft gate with a hard
threshold at the same operating point makes the trajectory diverge, as
too few cells remain to condition the normal matrix. This supports the
central design choice: weak cells should be treated neither as full
inliers nor removed categorically.

The remaining variants concern robustness rather than the main accuracy
gain. The default slope $\kappa=0.5$ outperforms the sharper and smoother
alternatives, and the result is insensitive to $\tau$ over a $2\times$
range. The single Cauchy pass is roughly neutral on these low-outlier
sequences; however, it protects the estimator in dynamic scenes, and the
three-component covariance, while only a modest APE\textsubscript{t} gain
here, prevents overconfidence along geometrically weak axes.

\subsection{Cross-Configuration Transfer}
\label{sec:results-selfcollected}

The self-collected sequences test whether the same front-end transfers
across hardware and motion regime. The dense front-end is run unchanged on
a TI AWR6843AOP-EVM, whose twelve-element virtual array provides a
$60^\circ/60^\circ$ azimuth/elevation field of view, rather than the
$60^\circ/15^\circ$ aperture of the AWR1843BOOST configurations used by the
public datasets. This is the only dataset with substantial elevation
coverage, and therefore the only one that exercises the vertical axis that
the per-cell elevation refinement of \cref{sec:wls} is designed to recover.
Despite the change in chip, aperture, and dominant motion, dense soft
weighting attains the lowest per-frame ego-velocity RMSE
($0.08$--$0.12$~\si{\meter\per\second}) on every sequence (\cref{tab:ate}),
confirming that the front-end transfers without retuning. The downstream
trajectory benefit of this improved velocity is examined next.

\subsection{From Ego-Velocity to Trajectory Accuracy}
\label{sec:results-coloradar}

\Cref{fig:trajectories} and \cref{tab:ate} report the fused-trajectory results
across the three evaluation datasets. On ColoRadar, the dense front-end
attains the lowest translational APE and RPE on all three sequences
(\cref{fig:trajectories}(a)--(d)). Mean APE\textsubscript{t} decreases from
$2.12$~m for the strongest point-cloud baseline, \ptclouddspcapon{}, to
$1.17$~m, corresponding to a \SI{45}{\percent} reduction. Mean
RPE\textsubscript{t} is also reduced to $0.92$~m, compared with $1.38$ and
$1.43$~m for the two point-cloud baselines.

Similar trends are observed on Radarize and the self-collected dataset. On the
Radarize sequences in \cref{fig:trajectories}(e)--(g), Dense Soft Weighting
achieves the lowest translational APE in \cref{tab:ate}, while the held-out
summary in \cref{fig:trajectories}(h) shows performance close to the Radarize
network and substantially better than milliEgo. On the self-collected rig
(\cref{fig:trajectories}(i)--(l)), the dense front-end again gives the lowest
translational APE on every sequence, reducing mean APE\textsubscript{t} from
$0.39$~m for the point-cloud baselines to $0.27$~m. Relative pose error is
more dataset-dependent: on the short \SI{1}{\meter} self-collected windows the
point-cloud baselines remain competitive, leading RPE\textsubscript{t} on two
of the three sequences (for example, $0.10$~m versus $0.16$~m on
\texttt{sequence\_01}), with dense soft weighting ahead only on
\texttt{sequence\_03}. These decimetre-level differences do not overturn the
APE gains; across point-cloud, learning-based, and cross-configuration
comparisons the fused trajectories show consistent improvements in global
translational accuracy.

Rotational errors are interpreted separately because the radar front-end is
fused as a velocity measurement in a loosely coupled ESKF. Attitude is
therefore governed primarily by IMU propagation, with radar affecting rotation
only indirectly through the filter state. This distinction is especially
important for Radarize, where rotational APE/RPE values are computed against a
consumer visual-inertial pseudo-ground truth while using the same device's
internal IMU as the inertial input. In contrast, ColoRadar and the
self-collected rig use dedicated higher-grade IMUs with independent lidar-SLAM
or motion-capture references. The larger rotational errors on Radarize
therefore mainly reflect the inertial and pseudo-ground-truth setup rather than
the dense radar weighting itself.

\subsection{Comparison Against Learning-Based Odometry}
\label{sec:results-radarize}

We evaluate dense soft weighting on the Radarize dataset (\cref{sec:datasets})
against the two learning-based baselines of \cref{sec:baselines}, the
Radarize Doppler-flow network and milliEgo. All methods are scored on the
same $89$-sequence held-out split under one protocol; the dense front-end
uses no training and keeps the fixed hyperparameters used throughout.

Over the full held-out split, the dense front-end achieves $1.11$~m mean
translational APE, compared with $4.22$~m for milliEgo and $0.81$~m for
Radarize (\cref{fig:trajectories}(h)). The Radarize network remains stronger in mean APE in its native
trained setting. At the median sequence, however, the gap narrows to
$0.77$~m versus $0.71$~m, and the dense front-end attains lower per-frame
forward-velocity RMSE on \SI{69}{\percent} of sequences. \Cref{tab:ate}
reports the three sequences plotted in \cref{fig:trajectories}(e)--(g), on
which dense soft weighting has the lowest translational APE. Together, these
results place the training-free dense front-end on par with, and in several
cases ahead of, learning-based odometry baselines, especially milliEgo, which
operates from sparse radar point-cloud data. Radarize provides a strong learned
dense baseline in a similar sensor and environment domain, but its performance
reflects a learned mapping for the Radarize distribution and may not transfer
directly when radar spectra change with sensor configuration or deployment
conditions. The computational trade-off is considered separately in
\cref{sec:runtime}, since deployment depends on both accuracy and per-frame
latency on embedded hardware.

\subsection{Runtime and Embedded Feasibility}
\label{sec:runtime}

We next evaluate this deployment cost by measuring per-frame latency
(\cref{tab:runtime}) on the two compute
platforms defined with the self-collected hardware in
\cref{sec:self-hardware}: a development laptop and a Jetson Orin~NX
embedded target. CPU figures use a single core
(\texttt{taskset~-c~0}, one BLAS thread); GPU figures use CUDA.

The cost of the proposed system front-end is governed by the range-Doppler cell
count listed in \cref{tab:runtime}: every cell undergoes the same beam
selection, angle refinement, and weighting, so single-core latency is
linear in the grid, at \SIrange{1.1}{1.2}{\micro\second} per cell from the
\num{3072}-cell Radarize cube (\SI{3.8}{\milli\second}) to the
\num{16384}-cell ColoRadar cube (\SI{17.8}{\milli\second}). For the larger
cubes GPU acceleration matters most, and the algorithm supports it
directly: the per-cell operations are independent, involve no detection
step or data-dependent branching, and the weighted least-squares estimate
reduces to an accumulation over cells, so the front-end maps onto
data-parallel hardware and runs in \SIrange{1.8}{2.1}{\milli\second}
almost independently of grid size. Every analytic front-end clears its
sensor budget (\SI{10}{\hertz} for ColoRadar and self-collected,
\SI{30}{\hertz} for Radarize) on one CPU core, on the laptop and on
Orin~NX. The learned baselines have a fixed inference cost set by the
input-map sizes; on the laptop GPU they run in \SI{3.4}{} (Radarize) and
\SI{1.2}{\milli\second} (milliEgo), but on Orin~NX both meet the
\SI{30}{\hertz} budget only on the integrated GPU.

\ptcloudonboard{} is excluded from \cref{tab:runtime} because its
detection stage runs on the radar's microcontroller rather than on the
host, so no end-to-end latency from the raw ADC stream can be measured on
a common compute target; its host-side step covers only the ego-velocity
estimator (\SIrange{0.5}{9}{\milli\second} across platforms), and the
on-chip cloud is unavailable on the self-collected rig.

\subsection{Discussion}
\label{sec:discussion}

Across the three datasets, two radar chips, and three radar
configurations evaluated above, a single fixed weighting rule improves
both per-frame velocity and trajectory accuracy without per-dataset
adjustment. This supports the representation-level claim of the present
work: the improvement arises from aggregating weighted evidence across
the dense spectrum, and because the comparison holds the ESKF back-end
fixed, it does not depend on back-end tuning.

The density of the measurement set also motivates the analytic outlier
handling that replaces RANSAC. RANSAC is well suited to sparse point
clouds with a high outlier fraction, where a hard inlier/outlier
partition over many randomly sampled minimal sets is required. On the
dense cube, the confidence weights already attenuate clutter and
sidelobe responses, and a single Cauchy reweighting suppresses the few
remaining outliers without discarding any cell. This pass is
deterministic and adds the cost of one additional weighted
least-squares evaluation, whereas the cost of RANSAC is stochastic and
depends on the hypothesis count. Because every cell is retained at a
graded weight, the covariance follows in closed form from the weighted
normal equations (\cref{eq:cov-wls}) rather than from a sandwich
estimate over an inlier set. Robust losses and iterative reweighting
are established in radar ego-velocity
estimation~\cite{Huang2024MRIO, RIVSLAM2024}; The proposed system applies the same
principle to the dense, soft-weighted cell set in a single pass rather
than to a pre-detected point cloud.

The improvement is most pronounced under lateral motion. Sparse
detections concentrate near boresight, which constrains the forward
velocity component but provides limited lateral geometry; the dense
front-end retains off-boresight cells at reduced weight and thereby
improves the conditioning of the $v_y$ estimate. The same analysis
identifies the conditions under which the benefit diminishes: when a
frame contains high-SNR detections with broad angular spread, binary
detection already yields a well-conditioned sparse measurement set and
the two front-ends produce similar estimates. Dense soft weighting is
therefore most beneficial under lateral or off-boresight motion, in
low-reflectivity scenes, and in single-axis geometries where binary
detection conditions the measurement set poorly.

Several limitations remain. The dominant residual error in the
self-collected sequences is bias rather than random noise: under
sidelobe leakage the per-cell angle-FFT direction estimate is
displaced toward boresight, so the unit vectors $\uvec{u}_i$
in~\cref{eq:wls} under-represent the lateral projection and bias
$\hat{v}_y$ during lateral motion. Weighted aggregation cannot remove
this error because it enters through the geometry matrix $\mat{U}$
rather than the residuals; a sidelobe-aware direction-of-arrival
estimator (Capon, MUSIC, or sparse Bayesian methods) is a principled
extension. The covariance model likewise assumes white residuals and
neglects angle-FFT sidelobe correlation across neighbouring bins, so it
may under-estimate uncertainty under heavy multipath. Finally, the
cross-platform evidence spans only single-chip devices at handheld and
cart speeds; aerial and higher-speed ground-vehicle settings remain to
be evaluated.

\section{Conclusion}
\label{sec:conclusion}

We presented \emph{Dense Soft Weighting}, an analytic radar front-end for sensing ego-velocity and covariance directly from dense range-Doppler spectra. By replacing binary detector output with continuous per-cell confidence, DSW preserves weak Doppler evidence while down-weighting ambiguous cells, enabling robust weighted least-squares velocity estimation and a closed-form, measurement-derived covariance for inertial fusion. Under a shared ESKF back-end, the proposed front-end reduces mean translational APE from 2.12 m to 1.17 m on ColoRadar and from 0.39 m to 0.27 m on the self-collected 3D handheld dataset. On Radarize, it closely matches or exceeds the accuracy of learned dense odometry methods while using the same analytic parameters and requiring no dataset-specific training. These results suggest that weak Doppler evidence across the radar field of view can contribute to ego-velocity estimation when retained with low confidence rather than removed by hard thresholding. This yields a radar front-end that transfers across different single-chip radar configurations while remaining lightweight enough for embedded deployment. The implementation runs in real time on embedded hardware, suggesting that further integration within on-chip mmWave radar processing pipelines may be feasible. Future work will investigate longer-horizon inertial smoothing back-ends that exploit the reported velocity covariance beyond per-frame updates, and additional validation on aerial and higher-speed platforms under more aggressive motion.

\bibliographystyle{unsrt}
\bibliography{references}

\end{document}